\documentclass[10pt,twocolumn,letterpaper]{article}

\usepackage{cvpr}              % To produce the CAMERA-READY version
\usepackage[accsupp]{axessibility}
\usepackage{graphicx}
\usepackage{amsmath}
\usepackage{amssymb}
\usepackage{booktabs}
\usepackage{bbm}
\usepackage[linesnumbered,ruled]{algorithm2e}
\usepackage{multirow}
\usepackage{booktabs,tabularx}
\usepackage{amsmath}
\usepackage{stfloats}
\usepackage[accsupp]{axessibility}  % Improves PDF readability for those with disabilities.

\usepackage[pagebackref,breaklinks,colorlinks]{hyperref}

\usepackage[capitalize]{cleveref}
\crefname{section}{Sec.}{Secs.}
\Crefname{section}{Section}{Sections}
\Crefname{table}{Table}{Tables}
\crefname{table}{Tab.}{Tabs.}

\def\cvprPaperID{8991} % *** Enter the CVPR Paper ID here
\def\confName{CVPR}
\def\confYear{2023}

\SetKwInput{KwInput}{Input}                % Set the Input
\SetKwInput{KwOutput}{Output}              % set the Output

\begin{document}

%%%%%%%%% TITLE - PLEASE UPDATE
% \title{Injecting Geometric Knowledge into Pre-training for \\Document Information Extraction}

% \title{GeoLayoutLM: Geometric Pre-training for Visually-Rich Document Understanding}

% \title{GeoLayoutLM: Geometric Pre-training for Relation Extraction in Visual Information Extraction}
\title{GeoLayoutLM: Geometric Pre-training for Visual Information Extraction}

\author{
Chuwei Luo$^*$, 
Changxu Cheng$^*$, 
Qi Zheng, 
Cong Yao \\
DAMO Academy, Alibaba Group\\
{\tt\small \{luochuwei,ccx0127,zhengqisjtu,yaocong2010\}@gmail.com}
% For a paper whose authors are all at the same institution,
% omit the following lines up until the closing ``}''.
% Additional authors and addresses can be added with ``\and'',
% just like the second author.
% To save space, use either the email address or home page, not both
% \and
% Second Author\\
% Institution2\\
% First line of institution2 address\\
% {\tt\small secondauthor@i2.org}
}
\maketitle
\def\thefootnote{*}\footnotetext{Both authors contributed equally to this work.}
\def\thefootnote{\arabic{footnote}}

%%%%%%%%% ABSTRACT
\begin{abstract}
  Visual information extraction (VIE) plays an important role in Document Intelligence. Generally, it is divided into two tasks: semantic entity recognition (SER) and relation extraction (RE). Recently, pre-trained models for documents have achieved substantial progress in VIE, particularly in SER. However, most of the existing models learn the geometric representation in an implicit way, which has been found insufficient for the RE task since geometric information is especially crucial for RE. Moreover, we reveal another factor that limits the performance of RE lies in the objective gap between the pre-training phase and the fine-tuning phase for RE. To tackle these issues, we propose in this paper a multi-modal framework, named GeoLayoutLM, for VIE. GeoLayoutLM explicitly models the geometric relations in pre-training, which we call geometric pre-training. Geometric pre-training is achieved by three specially designed geometry-related pre-training tasks. Additionally, novel relation heads, which are pre-trained by the geometric pre-training tasks and fine-tuned for RE, are elaborately designed to enrich and enhance the feature representation.
  According to extensive experiments on standard VIE benchmarks, GeoLayoutLM achieves highly competitive scores in the SER task and significantly outperforms the previous state-of-the-arts for RE (\eg, the F1 score of RE on FUNSD is boosted from 80.35\% to 89.45\%)
  \footnote{https://github.com/AlibabaResearch/AdvancedLiterateMachinery}.
\end{abstract}

%%%%%%%%% BODY TEXT
\section{Introduction}
\label{sec:intro}

Visual information extraction (VIE) is a critical part in Document AI\cite{cui2021document, Zhu2016SceneTD, Long2018SceneTD}. It has attracted more and more attention from both the academic and industrial community.
VIE involves semantic entity recognition (SER, \textit{a.k.a.} entity labeling) and relation extraction (RE, \textit{a.k.a.} entity linking) from visually-rich documents (VrDs) such as forms and receipts\cite{jaume2019funsd,zhang2021entity,cui2021document,li2021structext,xu2021layoutxlm, Shi2015AnET, Zhou2017EASTAE, wang2022multi}.
Recent years have witnessed the great power of pre-trained multi-modal models \cite{xu2020layoutlm,xu2021layoutxlm,xu2020layoutlmv2,huang2022layoutlmv3,li2021structurallm,li2021structext,li2021selfdoc,appalaraju2021docformer,gu2022unified,wang2022lilt,gu2022xylayoutlm,luo2022bivldoc,hong2022bros} in VIE tasks, especially the SER task.
Compared with SER, the RE task, which aims at predicting the relation between semantic entities in documents, has not been fully explored and remains a challenging problem\cite{li2021structext,hong2022bros}. RE is essential to provide additional structural information closer to human comprehension of the VrDs\cite{zhang2021entity}. It makes the open-layout information extraction possible, e.g., for open-layout key-value linking and form-like items grouping.

\begin{figure}[tp]
  \includegraphics[width=.99\linewidth]{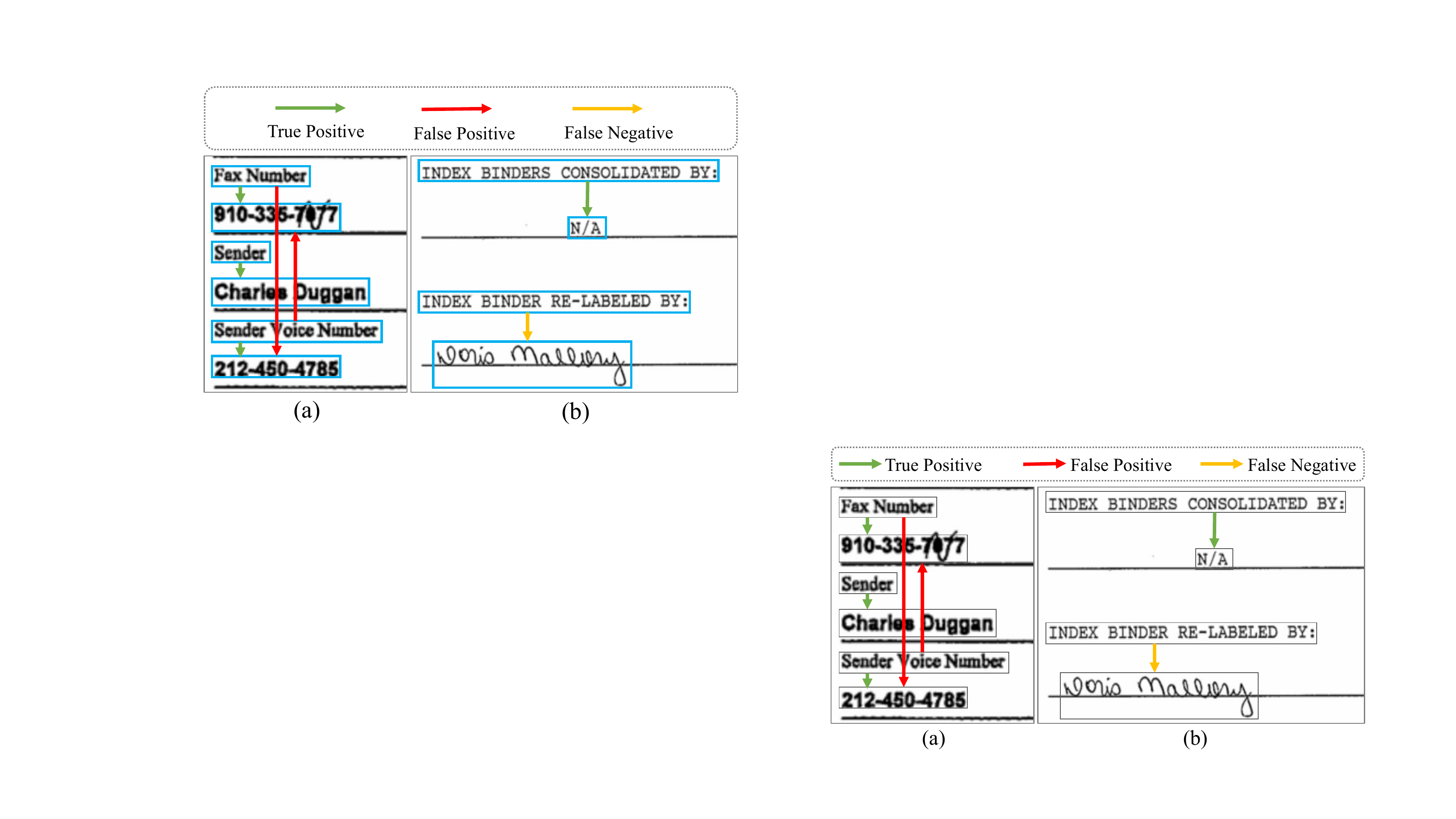}
  \caption{Incorrect relation predictions by the previous state-of-the-art model LayoutLMv3~\cite{huang2022layoutlmv3}. (a) LayoutLMv3 tends to link two entities relying more on their semantics than the geometric layout, \ie, the entity ``212-450-4785'' is linked to ``Fax Number'' regardless of their relationship in layout. (b) LayoutLMv3 successfully predicts the link in the upper half part but misses the link below, although both links are similar in geometric layout. These two examples clearly show \textbf{the importance of geometric information in relation extraction (RE)}.}
  \label{intro_case}
  \vspace{-2mm}
\end{figure}

\begin{table}[tp]
  \centering
  \begin{tabular}{lccc}
    \toprule[1pt]
    & \textbf{Precision} & \textbf{Recall} & \textbf{F1} \\
    \hline
    LayoutLMv3 & 75.82 & 85.45 & 80.35 \\
    + geometric constraint & \textbf{79.87} & 85.45 & \textbf{82.57} \\
    \bottomrule[1pt]
  \end{tabular}
  % \caption{The RE performance improvement by introducing a simple geometric restriction (on FUNSD). The precision gets rised remarkably while the recall is conserved since the false positive links are filtered carefully by geometric rules.}
  \caption{The RE performance improvement by introducing a simple geometric restriction (on the FUNSD dataset).}
  \vspace{-3mm}
  \label{intro_v3_geo}
\end{table}

% 聪哥
It is widely accepted that document layout understanding is crucial for VIE~\cite{xu2020layoutlm,xu2021layoutxlm,xu2020layoutlmv2,huang2022layoutlmv3,li2021structext,li2021selfdoc,appalaraju2021docformer,gu2022unified,wang2022lilt,gu2022xylayoutlm,luo2022bivldoc}, especially for RE~\cite{li2021structext,hong2022bros}.
The geometric relationships, a specific form for describing document layout, are important for document layout representations\cite{liu2019graph,luo2020merge,li2021structext}.
% Most works model the layout representation \textit{implicitly} by adding coordinates into input, introducing the relative position encoding or designing various pre-training tasks, including text-image alignment and masked image modeling.
Most previous pre-trained models for VrDs learn layout representations \textit{implicitly} by adding coordinates into the model inputs, combining the relative position encoding or supervising by alignment-related pre-training tasks like text-image alignment\cite{xu2020layoutlmv2,huang2022layoutlmv3,luo2022bivldoc} and masked vision language modeling\cite{xu2020layoutlm,xu2021layoutxlm,xu2020layoutlmv2,huang2022layoutlmv3,li2021structext,li2021selfdoc,appalaraju2021docformer,gu2022unified,li2021structext,hong2022bros}.
% However, it is not guaranteed that the geometric layout information is properly encoded after their feature extractor.
However, it is not guaranteed that the geometric layout information is well learned in these models.
Taking the state-of-the-art model LayoutLMv3 as an example, we find it would make mistakes in certain relatively simple scenarios, where the geometric relations between entities are not complicated.
% As shown in \cref{intro_case}, LayoutLMv3 seems to link two entities depending more on the semantics than the geometric layout, which may be caused by its weak representation for layout. 
As shown in \cref{intro_case}, LayoutLMv3 seems to link two entities depending more on the semantics than the geometric layout.
This indicates that its layout understanding is not sufficiently discriminative. To further verify our conjecture, we conduct an experiment by filtering the false positive relations using a simple geometric restriction (the linkings between entities should not point up beyond a certain distance), the precision would increase by a large margin (more than 4 points) while the recall is controlled unchanged, as detailed in \cref{intro_v3_geo}. This experiment proves that LayoutLMv3 does not fully exploit the useful geometric relationship information.
Besides, most existing methods did not directly take the relation modeling into consideration in pre-training.
They usually adopt token/segment-level classification or regression, which might underperform on downstream tasks related to relation modeling.
% As a result, it is urgent to have a better layout representation for RE task by learning the geometric relationship \textit{explicitly} in pre-training.
Therefore, it is necessary to learn a better layout representation for document pre-trained models by modeling the geometric relationships between entities \textit{explicitly} during pre-training.

During RE fine-tuning, previous works usually learn a task head like a single linear or bilinear layer\cite{li2021structext,hong2022bros} from scratch.
On the one hand, since the higher-level pair relationship features, which are beyond the token or text-segment features in documents, are complex, we argue that a single linear or bilinear layer is not always adequate to make full use of the encoded features for RE.
On the other hand, the RE task head initialized randomly is prone to overfitting with limited fine-tuning data.
Since the pre-trained backbone has shown tremendous potential \cite{devlin2018bert, dosovitskiy2020image}, why not pre-train the task head in some way simultaneously? Several works \cite{hu2022p, han2021adaptive,liu2022prompt} have proved that smaller \textit{gap} between pre-training and fine-tuning leads to better performance for downstream tasks. Hence, there is still considerable room for the design and usage of the RE task head.

Based on the above observations, we establish a multimodal pre-trained framework (termed as \textbf{GeoLayoutLM}) for VIE, in which a geometric pre-training strategy is designed to explicitly utilize the geometric relationships between text-segments, and elaborately-designed RE heads are introduced to mitigate the gap between pre-training and fine-tuning on the downstream relation extraction task.
Specifically, three geometric relations are defined: the relation between two text-segments (\textbf{GeoPair}), that among multiple text-segment pairs (\textbf{GeoMPair}), and that among three text-segments (\textbf{GeoTriplet}).
Correspondingly, three self-supervised pre-training tasks are proposed.
GeoPair relation is modeled by the \textbf{D}irection and \textbf{D}istance \textbf{M}odeling (\textbf{DDM}) task in which GeoLayoutLM needs to tell the direction of a directed pair and identify whether a segment is the nearest to another one in the direction.
Furthermore, we design a brand-new pre-training objective called \textbf{D}etection of \textbf{D}irection \textbf{E}xceptions (\textbf{DDE}) for GeoMPair, enabling our model to capture the common pattern of directions among segment pairs, enhance the pair feature representation and discover the detached ones.
For GeoTriplet, we propose a \textbf{C}ollinearity \textbf{I}dentification of \textbf{T}riplet (\textbf{CIT}) task to identify whether three segments are collinear, which takes a step forward to the modeling of multi-segments relations. It is important for non-local layout feature learning especially in form-like documents.
Additionally, novel relation heads are proposed to learn better relation features,
which are pre-trained by the geometric pre-training tasks to absorb prior knowledge about geometry, thus mitigating the gap between pre-training and fine-tuning.
Extensive experiments on five public benchmarks demonstrate the effectiveness of the proposed GeoLayoutLM.

Our contributions are summarized as follows:

\begin{itemize} \setlength{\itemsep}{0pt}
  \item[1)]  This paper introduces three geometric relations in different levels and designs three brand-new geometric pre-training tasks correspondingly for learning the geometric layout representation explicitly.  To the best of our knowledge, GeoLayoutLM is the first to explore the geometric relations of multi-pair and multi-segments in document pre-training.
  \item[2)]  Novel relation heads are proposed to benefit the relation modeling. Besides, the relation heads are pre-trained by the proposed geometric tasks and fine-tuned for RE, thus mitigating the object gap between pre-training and fine-tuning.
  % The novel relation heads are proposed to benefit the relation modeling in both geometric pre-training and RE fine-tuning. thus mitigating the object gap between pre-training and fine-tuning.
  % A novel Relation Feature Enhancement head is proposed to benefit the relation modeling. Besides, the heads for RE are pre-trained by the proposed geometric tasks, thus mitigating the object gap between pre-training and fine-tuning.
  \item[3)]  Experimental results on visual information extraction tasks including key-value linking as relation extraction, entity grouping as relation extraction, and semantic entity recognition show that the proposed GeoLayoutLM significantly outperforms previous state-of-the-arts with good interpretability. Moreover, our model has notable advantages in few-shot RE learning.
\end{itemize}

\section{Related Works}
\noindent\textbf{Visual information extraction}
Visual information extraction (VIE) aims at extracting entities from visually-rich document images, typically including semantic entity recognition (SER) and relation extraction (RE)\cite{jaume2019funsd,xu2022xfund,li2021structext,hong2022bros}.
Early works based on graph neural networks \cite{kipf2016semi,qian2018graphie,liu2019graph,qian2019graphie,luo2020merge,tang2021matchvie,yu2021pick} learned node features of text and layout in the downstream VIE tasks directly.
Recently, pre-training techniques have boosted the performance on document understanding. Various pre-training tasks are designed to learn better text/image features and their alignment for stronger multimodal document representation\cite{li2021structext,li2021selfdoc,appalaraju2021docformer,gu2022unified,wang2022lilt,gu2022xylayoutlm,lin2021vibertgrid,luo2022bivldoc,xu2020layoutlm,xu2021layoutxlm,xu2020layoutlmv2,huang2022layoutlmv3}.
Although they have achieved significant improvement on SER, RE remains largely underexplored and is also a challenging task\cite{zhang2021entity,hwang2020spatial,hong2022bros,li2021structext}.
% Zhang \etal adapt a dependency parsing model and a layout-aware encoder to the RE task.
% SPADE\cite{} propose an end-to-end framework that models complex spatial relationships in documents for both SER and RE.
% BROS\cite{hong2022bros} encoded the relative spatial positions of texts into BERT\cite{devlin2018bert} and propose a pre-training objective named area-masked LM to learn the layout representation better for SER and RE.
% BROS\cite{hong2022bros} encoded the relative spatial positions of texts into BERT\cite{devlin2018bert} and propose a pre-training objective named area-masked LM to learn the layout representation better for SER and RE.
BROS\cite{hong2022bros} encoded the relative spatial positions of texts into BERT\cite{devlin2018bert} to learn the layout representation better. In this paper, we focus on adopting pre-training to obtain better features.

\noindent\textbf{Geometric information} 
Geometric information is an important clue to represent the document layout.
% Intuitively, simply combining position features may help little with 2D semantic context, as the layout of the different documents varies a lot.
Liu \etal\cite{liu2019graph} utilized relative 2D positions in GNN.
GraphNEMR\cite{luo2020merge} incorporated the 8-geometry neighbours and geometry distance information in document modeling for SER.
SPADE\cite{hwang2020spatial} re-formulated the self-attention layer by introducing a relative spatial vector which is composed of relative coordinates, distance and angle embeddings. 
StrucText\cite{li2021structext} proposed a Paired Boxes Direction task to model the geometric direction of text-segments in pre-training.
However, these works only explored pair-level geometric relations. We expand geometric relation to more than two segments: the relations of multi-pairs and triplets are fully explored.

\noindent\textbf{Pre-training / Fine-tuning}
% Alleviating the gap between the pre-trained model and the downstream task have been the focus of recent studies in pre-trained models\cite{chronopoulou2019embarrassingly,howard2018universal,gururangan2020don,han2021adaptive,hu2022p}.
Recent studies on pre-training also focused on alleviating the gap between the pre-training stage and the downstream fine-tuning stage\cite{chronopoulou2019embarrassingly,howard2018universal,gururangan2020don,han2021adaptive,hu2022p,liu2022prompt}.
Hu \etal\cite{hu2022p} identified and studied the training schema gap and the task knowledge gap, and converted the downstream ranking task into a pre-training schema.
% Inspired by these works, we argue that during document pre-training, both relationship knowledge between text-segments and the relational model head structure are required for both document pre-training and RE task fine-tuning.
% Several pre-training techniques, such as continued pre-training\cite{howard2018universal,gururangan2020don}, using self-supervised tasks as an auxiliary task to help the target task\cite{chronopoulou2019embarrassingly}, are shown to be effective.
% \cite{hu2022p} achieve good performances in downstream ranking tasks by proposing a pre-trained model that consider the discrepancy between the knowledge needed in downstream ranking tasks and that learned during pre-training, and also mitigate the training schema gap regarding the differences in training objectives and model architectures.
Prompt-based models were proposed to adapt to various scenarios by converting downstream tasks to proper prompts which are consistent with the schema in pre-training\cite{liu2022prompt}.
Inspired by these works, we pre-train our elaborately-designed relation heads using the geometric tasks to absorb geometric knowledge adequately and improve its generalization in relation representation from the large-scale pre-training data.

%------------------------------------------------------------------------
% \section{Method}
\section{GeoLayoutLM}
GeoLayoutLM is a multi-modal framework for VIE. Geometric information is explicitly encoded and utilized by the novel geometry-based pre-training tasks and the pre-training of the elaborately-designed relation heads.
% \cref{model_arch} gives an overview of the GeoLayoutLM.
Additionally, an effective strategy for RE inference is introduced.

% \begin{figure*}
%   \centering
%   \includegraphics[width=0.99\linewidth]{./model_arch.pdf}
%   \caption{An overview of GeoLayoutLM. The left part is the backbone consisting of a vision module, a text-layout module and co-attention layers. The pre-training tasks shown in the top-right are two-folds, including the linguistic and the proposed novel geometric tasks. In the fine-tuning phase (bottom-right), the process of relation extraction is so similar to some geometric tasks that two pre-training heads (NPI, MDAD) are the same as two fine-tuning heads (CRP, RFE), which creates the condition for task head pre-training.}
%   \label{model_arch}
% \end{figure*}

\begin{figure}[tp]
  \centering
  \includegraphics[width=0.99\linewidth]{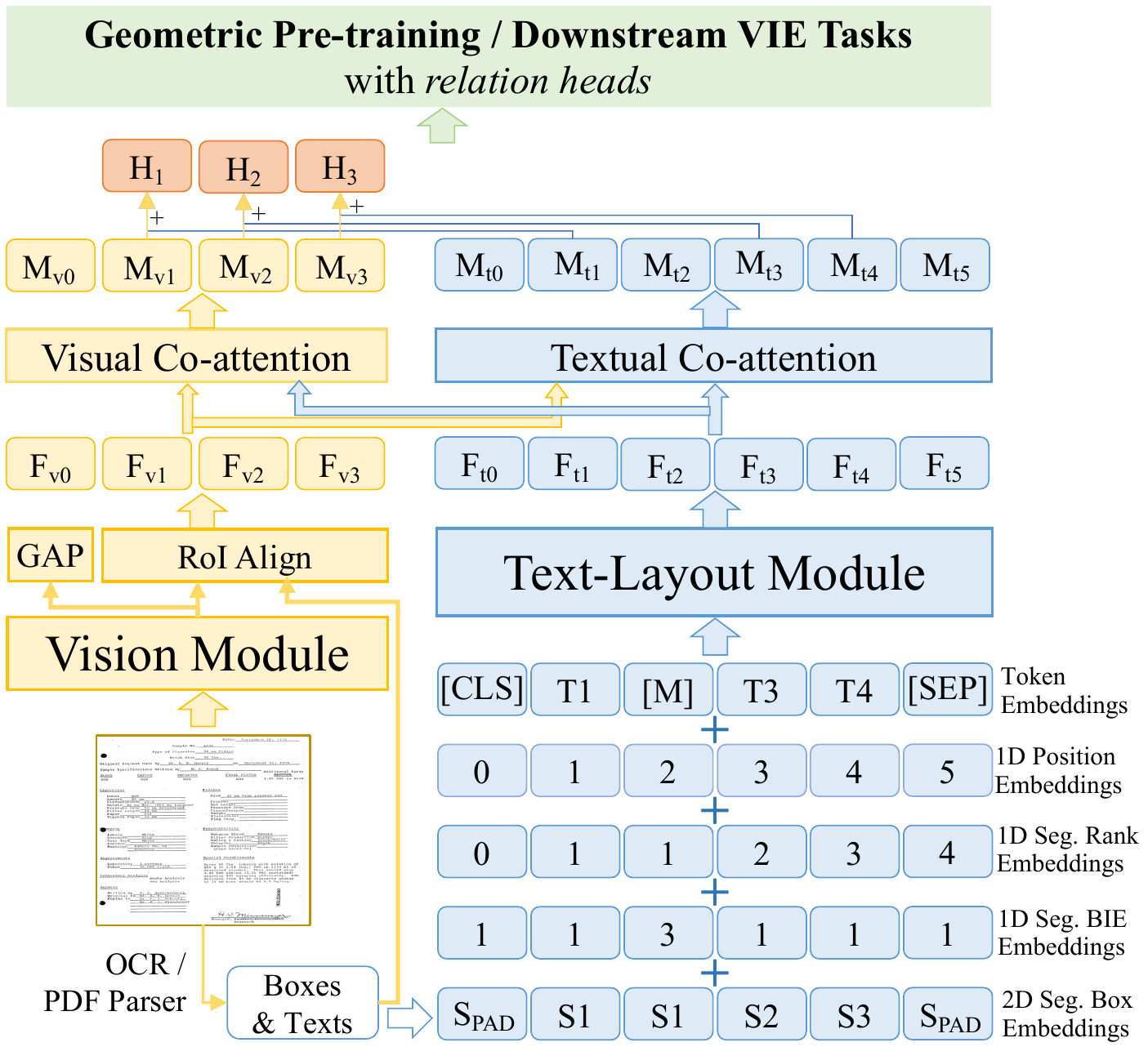}
  \caption{An overview of GeoLayoutLM.}
  \label{model_arch}
  \vspace{-0.20in}
\end{figure}

\subsection{Model Architecture}
% GeoLayoutLM has a vision-language model as the backbone feature extractor, and specific heads for various pre-training and fine-tuning tasks.

\subsubsection{Backbone}
\label{backbone_detail}

Inspired by the two-stream structure in METER\cite{dou2022empirical} and SelfDoc\cite{li2021selfdoc}, the backbone of GeoLayoutLM is composed of an independent vision module, a text-layout module, and interactive visual and text co-attention layers. As shown in \cref{model_arch}, the vision module takes the document image as input, and the text-layout module is fed with layout-related text embeddings.
Following LayoutLMv3\cite{huang2022layoutlmv3} and BiVLDoc\cite{luo2022bivldoc}, the text embeddings are the summation of 5 embeddings, including the token embeddings, 1D position embeddings, 1D segment rank embeddings, 1D segment BIE embeddings and 2D segment box embeddings.
The output feature of the vision module is processed by the global average pooling\cite{lin2013network} and RoI align\cite{he2017mask} to compute the global visual feature $F_{v0}$ and the $n$ visual segment features $\{F_{vi}|i\in [1,n]\}$.
Then the visual co-attention module takes $\{F_{vi}\}$ as query and $\{F_{ti}\}$ from the text-layout module as key and value for attention calculation, and outputs the fused visual features $\{M_{vi}\}$.
% Similarly, we can get $\{M_{ti}\}$ in the text side.
The fused textual features $\{M_{ti}\}$ are calculated in a similar way.
Finally, we add $M_{vi}$ and the corresponding first token feature of the segment $M_{t,b(i)}$ to obtain the $i$-th segment feature $H_i$.

\subsubsection{Relation Heads}
% Previous works focusing on document pre-training usually had a simple design on their downstream task head. For relation extraction, the final relation matrix was produced by a single linear or bilinear layer\cite{li2021structext,hong2022bros}. We argue that it is not enough for complicated tasks.

% 20221103 chuwei
The semantic entity recognition (SER) in VIE is usually modeled as a token classification problem.
Learning a simple MLP classifier is effective for SER\cite{huang2022layoutlmv3}.
In the relation extraction (RE) of VIE, the final relation matrix was usually produced by a single linear or bilinear layer\cite{li2021structext,hong2022bros}.
Since the relationships of text-segments are relatively complex and related to each other, we argue that a simple linear or bilinear layer is not enough for relation modeling.

\begin{figure}[tp]
  \centering
  \includegraphics[scale=0.8]{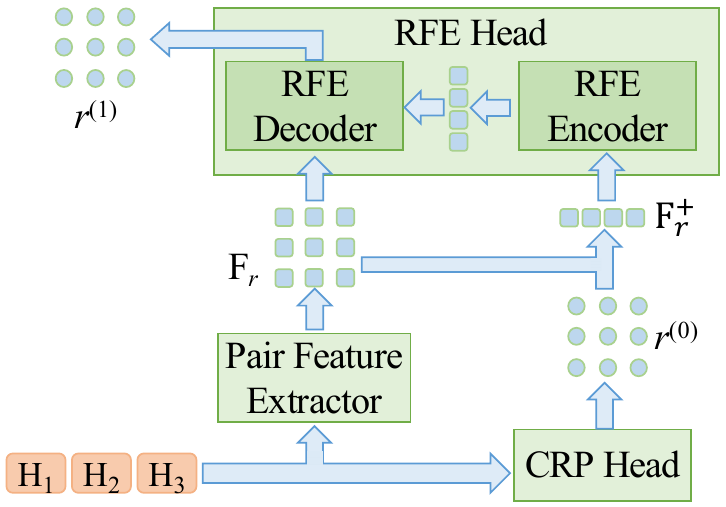}
  \caption{Relation heads.}
  \label{task_head_fig}
  \vspace{-3mm}
\end{figure}

In this work, we propose two relation heads, including a Coarse Relation Prediction (CPR) head and a novel Relation Feature Enhancement (RFE) head, to enhance the relation feature representation for both relation pre-training and RE fine-tuning.
% The RFE head is composed of a lightweight encoder-decoder transformer\cite{vaswani2017attention} and a fully-connected layer with sigmoid activation.
The RFE head is a lightweight transformer\cite{vaswani2017attention} consisting of a standard encoder layer, a modified decoder layer that discards the self-attention layer for computation efficiency, and a fully-connected layer followed by the sigmoid activation.
% As shown in \cref{task_head_fig}, given the segment features $\{H_i\}$, we first feed them to a bilinear layer to predict a coarse relation matrix $r^{(0)}$, which is a common operation we called Coarse Relation Extraction (CRE) here.
As shown in \cref{task_head_fig}, the text-segment features $\{H_i\}$ are fed into the CRP head (a bilinear layer) to predict a coarse relation matrix $r^{(0)}$.
To build the relation features $F_r$, the segment feature pairs are passed to a pair feature extractor (linearly mapping the concatenated paired features).
Then we select positive relation features $F_r^+$ based on $r^{(0)}$.
Note that $F_r^+$ probably has some false positive relation features since $r^{(0)}$ is the coarse prediction.
$F_r^+$ is then fed into the RFE encoder to capture the internal pattern of the true positive relations in each document sample, which is based on the assumption that most of the predicted positive pairs in $r^{(0)}$ are true.
All the relation features $F_r$ and the memory from the RFE encoder are fed into the RFE decoder to compute the final relation matrix $r^{(1)}$.
% In fact, the relation heads is a general design for relation modeling.

% The head for semantic entity recognition is a simple linear layer. We do not focus on this task in the work.

\subsection{Pre-training}
% We employ liguistic and geometric tasks to make our model learn the textual and layout representation respectively in the pre-training stage.

% As for the liguistic task, we simply adopt the widely-used Masked Language Model (MLM)\cite{devlin2018bert,xu2020layoutlm,hong2022bros} to learn contextually-rich text representation in token level. The supervision is exerted on both $\{M_{ti}\}$ and $\{F_{ti}\}$. 
% In this subsection, we will introduce the geometric tasks in detail.

% 20221029 chuwei
GeoLayoutLM is pre-trained with four self-supervised tasks simultaneously.
To learn multimodal contextual-aware text representations, the widely-used Masked Visual-Language Model (MVLM)\cite{xu2020layoutlm,xu2020layoutlmv2,huang2022layoutlmv3} is adopted on both $\{F_{ti}\}$ and $\{M_{ti}\}$.
Three proposed self-supervised geometric pre-training tasks are described in \cref{sec:geo_tasks}.

\subsubsection{Geometric Relationship}
To better represent document layout by geometric information, three geometric relationships are introduced, which are \textbf{GeoPair}, \textbf{GeoMPair} and \textbf{GeoTriplet}.
The relation between two text-segments (a pair) is denoted as \textbf{GeoPair}, which is also considered in previous works\cite{liu2019graph,luo2020merge,li2021structext} to model the relative layout information between two text-segments.
% A segment is just like a point in geometry, so a segment pair is like a line segment.
We further extend GeoPair to \textbf{GeoMPair} that is the relation among multiple segment pairs, to explore the relation of relations.
Like the relation of three points in geometry, \textbf{GeoTriplet} is also devised, which is the relation among three text-segments.

\subsubsection{Geometric Pre-training}
\label{sec:geo_tasks}
% We design three geometric tasks elaborately to learn layout representation and enable the pre-training of relation extraction head properly. The top-right part of \cref{model_arch} gives an overview. The three tasks are devised to cover three common geometric relations, which are that between two boxes (GeoPair), that among multiple paired-boxes (GeoMPair), and that among three boxes (GeoTriplet). The input of these tasks are segment features $\{B_i\}$ which can be either of the four features: $\{M_{vi}\},\{F_{vi}\},\{M_{t,b(i)}\},\{F_{t,b(i)}\}$, where $b(i)$ is the index of the first token of the $i$-th segment. A same pair feature extractor is used in both pre-training and fine-tuning.

% 20221031 chuwei
To make our model understand the geometric relationships and achieve good layout representations, we propose three geometry-related self-supervised pre-training tasks to model GeoPair, GeoMPair, and GeoTriplet respectively.
The input of these tasks are text-segments features $\{B_i\}$ which can be either of the five features: $\{H_{i}\},\{M_{vi}\},\{F_{vi}\},\{M_{t,b(i)}\},\{F_{t,b(i)}\}$, where $b(i)$ is the index of the first token of the $i$-th segment.

\begin{figure*}[tp]
  \centering
  \includegraphics[width=0.8\linewidth]{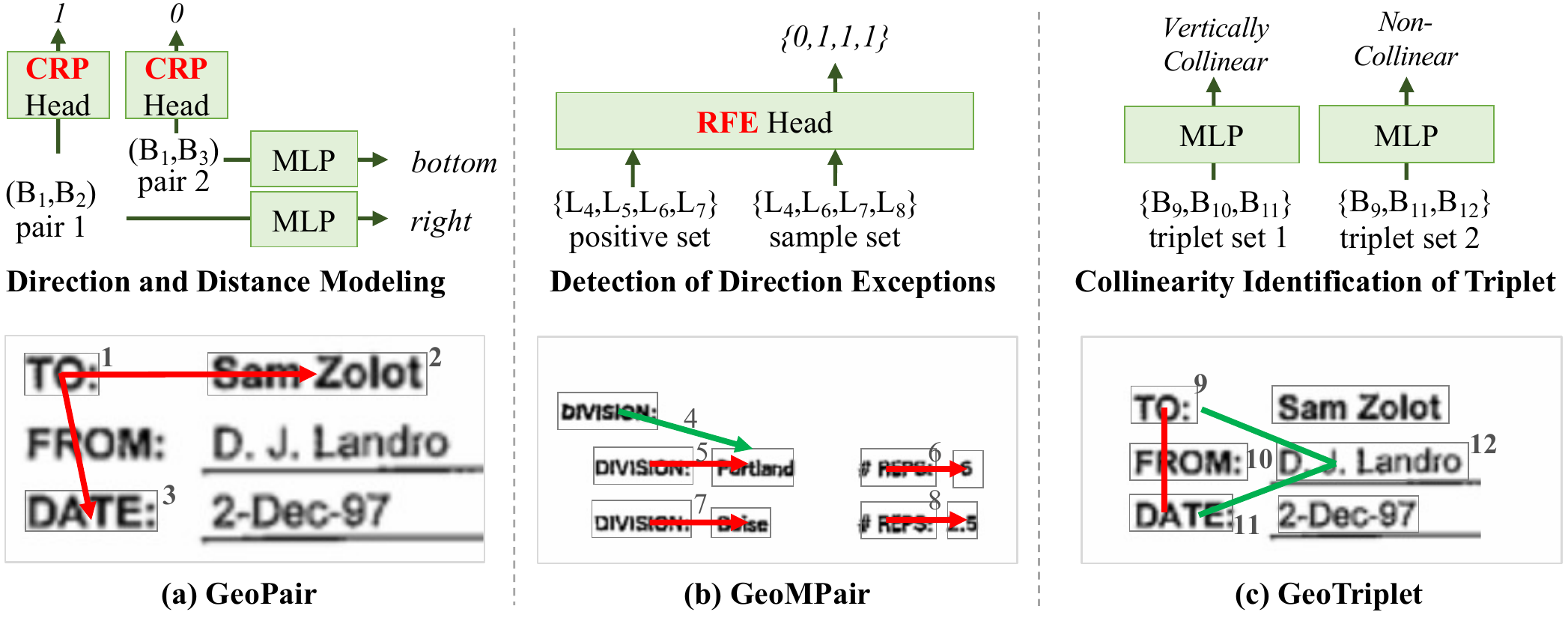}
  \vspace{-3mm}
  \caption{Geometric pre-training.}
  \label{geo_tasks_fig}
  \vspace{-6mm}
\end{figure*}

% \textbf{GeoPair}\ \ The relationship between two boxes can be measured by the direction and distance. A linear layer is taken as the Pair Direction Prediction (PDP) head to make 9-direction classification. As for distance, we define a Nearest Pair Identification (NPI) task instead of direct regression. There is at most 1 nearest box neighbor judged by distance in each of the 8 directions except the direction \textit{Overlapping}. The identification matrix is similar to the relation matrix in relation extraction, so the NPI shares the process of CRP, which achieves the goals of pre-training the CRP head.

% 20221031 chuwei
\noindent\textbf{Direction and Distance Modeling for GeoPair} To better understand the relative position relationship of two text-segments, as shown in \cref{geo_tasks_fig}(a), the Direction and Distance Modeling (DDM) is proposed, in which both the direction and distance are measured.

% Since the direction and distance of paired text-segments is from a OCR bbox to another OCR bbox, not the point to point direction and distance.
% As the text shapes in document are almost regular, the horizontal text bbox denoted by its top-left and bottom-right vertexes are adapted.

% \begin{figure}[t]
%   \centering
%    \includegraphics[width=0.9\linewidth]{./geo_def2.pdf}
%   %  \includegraphics[scale=0.6]{./geo_def2.pdf}
%    \caption{The basic definition on direction and distance of two lines or boxes. The anchor line and box are denoted in orange. (a) A case in a single dimension (the horizontal here; the vertical case is similar). (b) A case in real document. There are 8 directions except \textit{Overlapping} here.}
%    \label{geo_def}
% \end{figure}

% For the direction $\mathrm{D}$ from a box $\mathbf{b}_1$ to another box $\mathbf{b}_2$, motivated by \cite{luo2020merge}, a compound of two orthogonally-dimensional directions is introduced.
% In the horizontal dimension, there are 3 directions from a line segment to another, \ie, left, overlap and right.
% Similarly, the 3 directions in the vertical dimensions are top, overlap and bottom.
% So the direction $\mathrm{D}$ has 9 possible values, as shown in \cref{geo_def}(b). Compared with the direction definition in StrucTexT\cite{li2021structext}, ours is more concrete for 2D boxes with various aspect ratios.
We consider 9 directions, including 8 neighbor ones\cite{luo2020merge} and the overlapping.
Hence, the direction modeling is exactly a 9-direction classification problem:

\begin{equation}
  P_{ij}^{direct} = \mathrm{Softmax}(\mathrm{Linear}([B_i, B_j]))
  \label{dierecion}
\end{equation}
where $P_{ij}^{direct}$ is the predicted direction probability, $[\cdot]$ is the concatenation operation.

% As for the distance modeling, Euclidean distance is used to measure the distance between two bounding boxes of paired text-segments.
% The 1-d distance between two segments is defined as the minimum distance between the points in each line segment.
% The overlapped pairs are at a distance of 0.
% \cref{geo_def}(a) shows some examples on how to measure the distance.
The distance between two segments is defined as the minimum distance between the two bounding boxes\cite{luo2020merge}.
The distance is modeled as a binary classification problem that is to identify whether the $j$-th text-segment is the nearest to the $i$-th one in their direction. There is at most 1 nearest segment judged by distance in each of the 8 neighbor directions.
A bilinear layer is applied here:
\begin{equation}
  P_{ij}^{dist} = \mathrm{Sigmoid}(\mathrm{Bilinear}(B_i, B_j))
  \label{distance}
\end{equation}
where $P_{ij}^{dist}$ is the probability of the nearest pair identification.
Note that the operation in \cref{distance} shares the same process of CRP, which achieves the goals of pre-training the CRP head.

The loss function $\mathcal{L}_{DDM}$ of DDM task is defined as:
\vspace{-2mm}
\begin{equation}
  \begin{aligned}
  \mathcal{L}_{DDM} &= \mathrm{CrossEntropy}(P^{direct},Y^{direct}) \\
  &+ \mathrm{BCELoss}(P^{dist},Y^{dist})
  \end{aligned}
  \label{loss_ddm}
\end{equation}
where $Y^{direct}$ and $Y^{dist}$ are labels for direction and distance modeling.

% \textbf{GeoMPair}\ \ To explore the relation among pairs, we devise a Multi-pair Direction Anamaly Detection (MDAD) task, in which we need to discriminate pairs whether their directions are anomalous in the \textsf{sample set}. A direction is regarded anomalous if its pairs have a minor ratio in the given \textsf{positive set}. For example, given a few segment pairs as the positive set in which more than 60\% have a same direction \textit{Right}, and an arbitrary sample set, the right pairs in the sample set are labeled as 1, while the pairs of other directions are labeled as 0 (anomaly), as shown in \cref{model_arch}. We first build the pair features $\{P_{ij}\}$ by the pair feature extractor, then feed the positive set and the sample set to the MDAD head. Note that the MDAD head has the same architecture as RFE head, thus sharing the same process, which also makes it possible to pre-train the RFE head.

% 20221101 chuwei
% \noindent\textbf{Detection of Direction Exceptions for GeoMPair} Intuitively, it is reasonable to believe the GeoMPair relation of multiple paired text-segments can well represent the overall layout of the corresponding region in the document.
% As shown in \cref{geo_tasks_fig} (b), multiple paired text-segments with the same direction in the document indicate that this is a wireless form area in this document.
\noindent\textbf{Detection of Direction Exceptions for GeoMPair} The relationships within a certain document area usually have some common geometric attributes. As shown in \cref{geo_tasks_fig}(b), the directions of key-value pairs are the same (arrows in red) in the wireless form area. The link in green can be easily judged as false due to its exceptional direction.
Motivated by this, the Detection of Direction Exceptions (DDE) task is proposed to model GeoMPair for the non-local layout understanding in documents.

% The DDE task is to predict the paired text-segments direction exceptions in a \textsf{sample set} $S = {s_1,..., s_n}$.
The DDE task is to discriminate segment pairs whether their directions are exceptional in a \textsf{sample set} $S$.
% The direction of a segment pair is regarded as an exception if it is not the main direction in the given \textsf{positive set} $S_p$ ($S_p\in S$).
A direction is regarded as an exception if the pairs with the direction have a minor ratio in the given \textsf{positive set} $S_p$.
For example, given a positive set in which more than 60\% of pairs have the same direction \textit{Right}, and a sample set, we label the right pairs in the sample set as 1, while the pairs of other directions as 0 (exception), as shown in \cref{geo_tasks_fig}(b).
The pair feature $L_i$ is built by linearly projecting the concatenated segment features.
Then, the positive set and the sample set are fed into the proposed RFE head to predict the discrimination probabilities of the sample set. The binary cross-entropy loss is applied.
\vspace{-2mm}

\begin{align}
  &P^{DDE} = \mathrm{RFE}(S_p,S) \\
  &\mathcal{L}_{DDE} = \mathrm{BCELoss}(P^{DDE}, Y^{DDE})
  \label{rfe_out}
\end{align}

% \textbf{GeoTriplet}\ \ Like the collinear attribute of three points, we define that of three boxes to take a step forward on the road of learning multi-elements layout. Given three boxes A, B and C, if the directions from A to B, from B to C and A to C are the same or opposite, they are collinear; otherwise non-collinear. The collinear cases can be further divided into four classes: horizontal line, vertical line, forward slash and backslash. During pre-training, the triplet feature is the summation of the three segment features since the collinear attribute is undirected. Then we feed it to a linear layer to make the 5-classification.

% 20221101 chuwei
\noindent\textbf{Collinearity Identification of Triplet for GeoTriplet} 
The geometric alignment of segments is an important expression of document layout, which is meaningful and involves the relation of multiple segments. 
Like the collinear attribute of three points, we define that of three segments. 
Given three text-segments $B_i$, $B_j$ and $B_k$, if the direction from $B_i$ to $B_j$, $B_j$ to $B_k$ and $B_i$ to $B_k$ are the same or antiphase, they are collinear; otherwise non-collinear. The collinear cases can be further divided into four classes: horizontal line, vertical line, forward slash and backslash.
As shown in \cref{geo_tasks_fig}(c), the left-aligned segments with the same entity tag are vertically collinear.
Correspondingly, a pre-training task called Collinearity Identification of Triplet (CIT) is proposed.
% the geometric collinearity of the text-segments collinearity is a good representation of layout alignment.
% Since at least three text-segments can express collinearity in a document, a new collinearity identification of triplet (CIT) pre-training task is introduced for the GeoTriplet relations.
% The calculation of direction from $B_i$ to $B_j$ is the same in DDM.
The triplet feature is the summation of three segment features since the collinear attribute is undirected. Subsequently, the 5-classification is made for CIT:
\vspace{-2mm}
\begin{align}
  &P^{CIT}_{ijk} = \mathrm{Softmax}(\mathrm{Linear}(B_i+B_j+B_k)) \\
  &\mathcal{L}_{CIT} = \mathrm{CrossEntropy}(P^{CIT}, Y^{CIT})
  \label{cit_loss}
\end{align}
% where $P^t_{ijk}$ represents the predicted collinear probability distribution. $Y^t_{ijk}$ is the collinear label of the three text-segments $B_i$, $B_j$ and $B_k$.

The full pre-training objective of GeoLayoutLM is:
\begin{equation}
  \mathcal{L}_{pt} = \mathcal{L}_{MVLM} + \mathcal{L}_{DDM} + \mathcal{L}_{DDE} + \mathcal{L}_{CIT}
  \label{full_loss}
\end{equation}

\subsection{Fine-tuning and Inference}
% \subsection{Fine-tuning GeoLayoutLM for VIE}
% In fine-tuning, all the modules are initialized with the pre-training ones except the token prediction head.
During fine-tuning, the relation heads are initialized with the pre-trained parameters, which mitigates the gap between pre-training and fine-tuning.
The cross-entropy and binary cross-entropy function are utilized in SER loss $\mathcal{L}_{\mathrm{SER}}$ and RE losses $\{\mathcal{L}_{\mathrm{RE}, i}|i=0,1\}$ (corresponds to $r^{(0)}$ and $r^{(1)}$) respectively.
They are optimized together:
\vspace{-2mm}
\begin{equation}
  \mathcal{L}_{ft}=\mathcal{L}_{\mathrm{SER}}+\sum _{i=0}^1 \mathcal{L}_{\mathrm{RE},i}
  \label{loss_ft}
\end{equation}

In the RE task, for a relation pair $B_{j}\to B_{i}$, $B_{j}$ is called the father node, and $B_{i}$ is the son node.
$r_{i,j}^{(1)}$ stands for the probability that $B_{j}$ is the father of $B_{i}$.
The final relation output $\mathbb{R}$ was usually defined as: $\mathbb{R}_{ij}=\mathbbm{1}\left(r_{i,j}^{(1)}>0.5\right)$, where $\mathbbm{1}(\cdot)$ is the indicator function.
Optionally, we propose to impose the Restriction on the Selection of Fathers (RSF) for each son during the inference if some segments have several father nodes.
Specifically, the $j$-th segment is regarded as a father node of the $i$-th one only if $r_{i,j}^{(1)}>0.5$ and $r_{i,j}^{(1)}$ is close to the maximum probability:
% \vspace{-2mm}

% Additionally, a Restriction on the Selection of Fathers (RSF) mechanism is introduced for each son during the inference if some segments have several father nodes. 
% Specifically, the $B_j$ is regarded as a father node of $B_i$ only if $r_{i,j}^{(1)}>0.5$ and $r_{i,j}^{(1)}$ is close to the maximum probability:

\begin{equation}
  \mathbb{R}_{ij}=\mathbbm{1}\left(r_{i,j}^{(1)}>0.5\right)\times \mathbbm{1}\left(\max _k r_{i,k}^{(1)}<r_{i,j}^{(1)}+\tau\right)
  \label{infer_1}
\end{equation}

% where $\mathbbm{1}(\cdot)$ is the indicator function, and $\tau$ is the threshold measuring how close between the probabilities. In this paper, $\tau$ is $1e-3$.
\noindent{where} $\tau$ is the margin between the probabilities.

% threshold measuring how close to the maximum probability is ok.

We suggest an additional variance loss for RSF especially. Since the probabilities of father nodes are expected to be as close as possible, the variance of them should be as small as possible. During fine-tuning, the variance loss is exerted on the son nodes that have more than 1 father node.

\section{Experiments}
\subsection{Implementation Details}
The backbone detail is described in \cref{backbone_detail}. 
The vision module is composed of a ConvNeXt\cite{liu2022convnet} and a multi-scale FPN\cite{liao2020real}. BROS\cite{hong2022bros} is used as our text-layout module. The visual and textual co-attention modules are both equipped with a transformer decoder layer.

% In all our experiments, the text-layout module is BROS-large\cite{hong2022bros} due to its impressive way to encode the 2D relative position.
% For the visual part in the GeoLayoutLM, we use the backbone of a ConvNeXt
The document images are resized to $768\times 768$. The embedding size and feed-forward size of the co-attention module are 1024 and 4096 respectively.
% The maximum of the token sequence length and the segment number is set to 512 and 256 respectively.
In the RFE head, the relation feature size and feed-forward size are both set to 1024. The number of attention heads is 2. $\tau$ is $1e-3$.

% \subsubsection{Pre-training Details}
\noindent \textbf{Pre-training Details.}
Following the LayoutLM series\cite{xu2020layoutlm,xu2020layoutlmv2,huang2022layoutlmv3}, we pre-train our model on the IIT-CDIP Test Collection 1.0\cite{lewis2006building}, which consists of around 11 million document images.
However, the original line-level OCR annotation in the dataset will lead to the monotony of paired box directions (only top and bottom exist in most documents), which is extremely harmful to geometric pre-training.
To break the imbalance in the distribution of the geometric relationship, we modify the OCR annotation by a Poisson Line Segmentation algorithm for each document with a probability of 90\%. \cref{alg_seg} lists the procedure of splitting a line.

\begin{algorithm}
  \caption{Poisson Line Segmentation}\label{alg_seg}
  \KwInput{An original line from OCR annotation $L$}
  \KwOutput{The processed line(s) $L'$}
  Get the number of words in $L_i \to N_w$\;
  $p_l = (1 - 1 / (N_w - 0.5))$; \footnotesize{\tcp{split probability}}
  \eIf{$N_w < 2 || \mathrm{rand()} > p_l$} {
    $L' \gets L$
  }{
    $N_s = \mathbf{poisson}(\lambda =\min(N_w/3, 7))$\;
    Split $L$ into $N_s$ segments $\to L'$\;
  }
\end{algorithm}

The AdamW optimizer is applied for pre-training, with the initial learning rate of 1e-5 and a linear decay learning rate scheduler.
% The GeoLayoutLM is trained by using the Adam optimizer with the learning rate of 1e-5 and a linear decay learning rate schedule.
We use a batch size of 224 to train GeoLayoutLM for 2 epochs on the IIT-CDIP dataset.
The maximum sequence length is set to 512. The maximum number of text-segments is set to 256.
Following \cite{xu2020layoutlm,xu2020layoutlmv2,huang2022layoutlmv3}, we mask 15\% text tokens in which 80\% are replaced by the [MASK] token, 10\% are replaced by a random token, and 10\% keeps unchanged.
In DDM, 16 text-segments are randomly sampled. Then, for each of them, we randomly sample 32 different text-segments to build paired text-segments.
In DDE, 40 segment pairs are randomly sampled in the document.
In CIT, 16 triple text-segments are randomly sampled.

% \subsubsection{Fine-tuning Details}
\noindent \textbf{Fine-tuning Details.}
% For the downstream visual information extraction task, especially in extract relation, we conduct fine-tuning on two popular benchmark datasets. \textbf{FUNSD}\cite{jaume2019funsd} is a scanned document dataset for form understanding. It has 149 training samples and 50 test samples with multifarious layout. We focus on both the semantic entity recognition (a.k.a. entity labeling) and the relation extraction (a.k.a. entity linking) tasks for the complete goal of information extraction in open-layout document. \textbf{CORD}\cite{park2019cord} is a camera-captured receipt dataset for key information extraction. It contains 800 training, 100 validation and 100 test images. In CORD, the RE capability can be applied to two aspects: entity grouping and key-value discrimination.
% 20221103 chuwei
Following the LayoutLM series\cite{gu2022xylayoutlm,xu2020layoutlmv2,huang2022layoutlmv3}, the SER task is regarded as a sequence labeling problem aiming to tag each word with a label.
For the RE task, to conduct fair comparisons with previous methods (e.g., BROS~\cite{hong2022bros}), the ground truth entity labels are used.
We evaluate GeoLayoutLM on two popular benchmark datasets with five subtasks.
\textbf{FUNSD}\cite{jaume2019funsd} is a scanned document dataset for form understanding. It has 149 training samples and 50 test samples with multifarious layouts. We focus on both the semantic entity recognition (a.k.a. entity labeling) and the relation extraction (a.k.a. entity linking) tasks.
\textbf{CORD}\cite{park2019cord} is a camera-captured receipt dataset for information extraction. It contains 800 training, 100 validation and 100 test images. In CORD, three subtasks are evaluated including semantic entity recognition (SER), relation extraction as entity grouping (REaGRP) and relation extraction as key-value linking (REaKV).
We fine-tune our GeoLayoutLM for 200 epochs in FUNSD and 100 epochs in CORD with the batch size of 6. The learning rate is initially set to $2e-5$.

\subsection{Comparison with the SOTAs}
We compare our results with the previous state-of-the-arts. As shown in \cref{res_sota}, our GeoLayoutLM obtains the best F1 score in both semantic entity recognition (SER) and relation extraction (RE). 

\begin{table*}[tp]
  \centering
  \begin{tabular}{m{.45\columnwidth}|m{.20\columnwidth}|m{.16\columnwidth}<{\centering}|m{.16\columnwidth}<{\centering}|m{.16\columnwidth}<{\centering}|m{.16\columnwidth}<{\centering}|m{.16\columnwidth}<{\centering}}
    \toprule[1pt]
    \multirow{2}*{\textbf{Method}} & \multirow{2}*{\#\textbf{Params}} & \multicolumn{2}{c|}{\textbf{FUNSD}} & \multicolumn{3}{c}{\textbf{CORD}}\\
    \cline{3-7}
     & & SER & RE & SER & REaKV & REaGRP\\
    \hline
    BERT$_{LARGE}$\cite{devlin2018bert} & 340M & 65.63 & 29.11 & 90.25 & - & - \\
    LayoutLM$_{LARGE}$\cite{xu2020layoutlm} & 343M & 78.95 & 42.83 & 94.93 & - & - \\
    StrucTexT\cite{li2021structext} & 107M & 83.09 & 44.10 & - & - & -\\
    SERA\cite{zhang2021entity} & - & - & 65.96 & - & - & -\\
    LayoutLMv2$_{LARGE}$\cite{xu2020layoutlmv2} & 426M & 84.20 & 70.57 & 96.01 & - & 97.29\\
    BROS$_{LARGE}$\cite{hong2022bros} & 340M & 84.52 & 77.01 & 97.28 & - & 97.40\\
    LayoutLMv3$_{LARGE}$\cite{huang2022layoutlmv3} & 357M & 92.08 & 80.35$\dagger$ & 97.46 & 99.64$\dagger$ & 98.28$\dagger$\\
    \hline
    GeoLayoutLM & 399M & \textbf{92.86} & \textbf{89.45} & \textbf{97.97} & \textbf{100.00} & \textbf{99.45} \\
    \bottomrule[1pt]
  \end{tabular}
  % \caption{Comparison with related methods. The F1 score followed by * means it is re-implemented by BROS, and that followed by $\dagger$ is re-implemented by us. In CORD, we conduct relation extraction as both key-value link and entity grouping. All the listed methods are the large models except that the StrucTexT is a base model.}
  \caption{Comparison with existing models that explore both SER \& RE. The F1 score followed by $\dagger$ means it is re-implemented by us.}
  \label{res_sota}
  \vspace{-3mm}
\end{table*}

For the FUNSD SER task, GeoLayoutLM and LayoutLMv3 both significantly surpass other models.
Besides, the SER results on FUNSD and CORD also suggest that the geometric pre-training does the SER slightly more favorable than the popular text-image alignment. 
For the RE task, GeoLayoutLM significantly outperforms the previous state-of-the-art by 9.1\% on FUNSD, and reaches or nearly reaches the perfect performance in CORD. It demonstrates the great superiority of our model in extracting relations.
Even if we only fine-tune for 100 epochs, we still achieve (SER: 92.24\%, RE: 88.80\%) on FUNSD.

% GeoLayoutLM is slightly heavy due to the combination of independent vision and text-layout modules.
GeoLayoutLM backbone is slightly heavy due to the two-tower encoder.
Our vision module is flexible and can be replaced by others.
LayoutLMv3 has a coupling feature encoder for visual patches and text, which contributes to fewer parameters.
% However, they may struggle to encode the fine-grained visual representation.
The relation head we used in LayoutLMv3 is the same as the CPR head in GeoLayoutLM (1M Params).
The proposed RFE head (14M) only constitutes 3.5\% of the total parameters.
On one Nvidia V100 GPU, the average inference time of GeoLayoutLM is 80.17ms, which is nearly the same as that of LayoutLMv3 (79.69ms).

\subsection{Ablation Study}
% We make plentiful ablation studies to verify the effectiveness of geometric pre-training, the design of the RE heads and the RSF strategy respectively.
To better understand the effectiveness of geometric pre-training, the design of the RE heads and the RSF strategy in GeoLayoutLM, we perform plentiful ablation studies.

% \subsubsection{Impact of Geometric Pre-training}
\noindent \textbf{Impact of Geometric Pre-training.}
To figure out how each pre-training task influences the information extraction result, we pre-train our model using different combinations of the geometric tasks while remaining the MLM task. To be efficient, only 10\% of the original pre-training data is used to train the model for 1 epoch.
The results meet our expectations completely, as shown in \cref{abl_geo}. By comparing \#0 with \#1x, we observe that the performance of SER and RE will be improved if either of the three geometric tasks is exerted paralleled to the MLM task. For SER, GeoPair contributes the most while GeoMPair does the least. For RE, GeoMPair contributes the most while GeoTriplet does the least, which may be owing to the RE head that is directly pre-trained in GeoPair and GeoMPair. By comparing \#2x with \#1x and \#3x with \#2x, we find that it is always better to pre-train with more geometric tasks, which indicates that the tasks are complementary.

\begin{table}
  \centering
  \begin{tabular}{m{.05\columnwidth}<{\centering}|m{.11\columnwidth}<{\centering}m{.14\columnwidth}<{\centering}m{.17\columnwidth}<{\centering}|m{.11\columnwidth}<{\centering}m{.11\columnwidth}<{\centering}}
    \toprule[1pt]
    \# & GeoPair & GeoMPair & GeoTriplet & SER & RE\\
    \hline
    0 &&&& 83.39 & 74.91\\
    1a & \checkmark &&& 91.80 & 82.23\\
    1b && \checkmark && 88.67 & 82.56\\
    1c &&& \checkmark & 90.78 & 78.90\\
    \hline
    2a & \checkmark & \checkmark && 91.86 & 85.22\\
    2b & \checkmark && \checkmark & 91.90 & 82.37\\
    2c && \checkmark & \checkmark & 91.39 & 84.97\\
    \hline
    3 & \checkmark & \checkmark & \checkmark & 92.17 & 85.32\\
    \bottomrule[1pt]
  \end{tabular}
  \caption{Ablation study on the geometric pre-training task in FUNSD. The first column labels the experiment settings.}
  \label{abl_geo}
  \vspace{-2mm}
\end{table}

\begin{table}[tp]
  \centering
  \begin{tabular}{lccc}
    \toprule[1pt]
    & \textbf{Entropy}$\downarrow$ & \textbf{Cross Entropy}$\downarrow$ & \textbf{Acc.}$\uparrow$ \\
    \hline
    LayoutLMv3 & 1.1423 & 0.9986 & 0.6319 \\
    GeoLayoutLM    & \textbf{0.7633} & \textbf{0.5884} & \textbf{0.8223} \\
    \bottomrule[1pt]
  \end{tabular}
  \caption{Experiments on the geometric layout understandings. The entropy of direction prediction reveals the information maintained in the backbone. The lower the Entropy and the Cross Entropy are, the more layout information the model maintains.}
  \label{tab:geo_rel_usage_intro}
  \vspace{-4mm}
\end{table}

To be interpretable,
we also investigate how much information of geometric relationship is kept after an example is encoded for the downstream information extraction task. To this end, we exert a linear classifier onto the backbone of the model fine-tuned on FUNSD (GeoLayoutLM VS LayoutLMv3), and \textit{only train the classifier} on the re-processed FUNSD dataset with pair direction labels (9-direction classification), to squeeze the direction information that is measured by the classification entropy, cross-entropy and the accuracy. As shown in \cref{tab:geo_rel_usage_intro}, GeoLayoutLM has a lower entropy and cross-entropy, and a higher accuracy, indicating that it retains much more information about geometric relations in the downstream tasks.

\begin{figure}[tp]
  \includegraphics[scale=0.38]{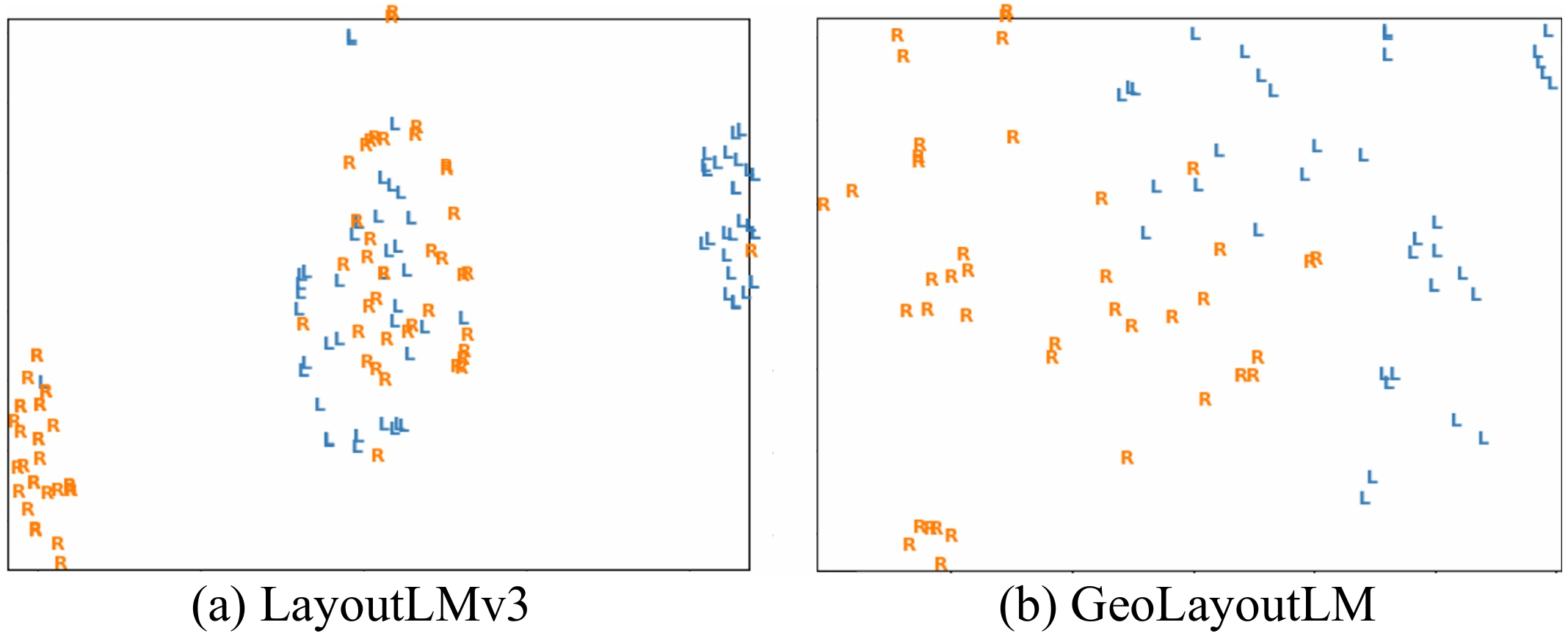}
  \caption{Comparison of the left (L) and right (R) relation features. For LayoutLMv3 and GeoLayoutLM, 2D layout positions remain unchanged and the input text tokens are set to [UNK].}
  \label{fig:vis_emb_intro}
  \vspace{-4mm}
\end{figure}

To further understand the geometric layout information, we conduct an embedding visualization of the left-right direction. As shown in \cref{fig:vis_emb_intro}, GeoLayoutLM has stronger distinguishable embeddings in left and right relationships that are important for document layout representation.

\begin{figure}[t]
  \centering
   \includegraphics[width=0.99\linewidth]{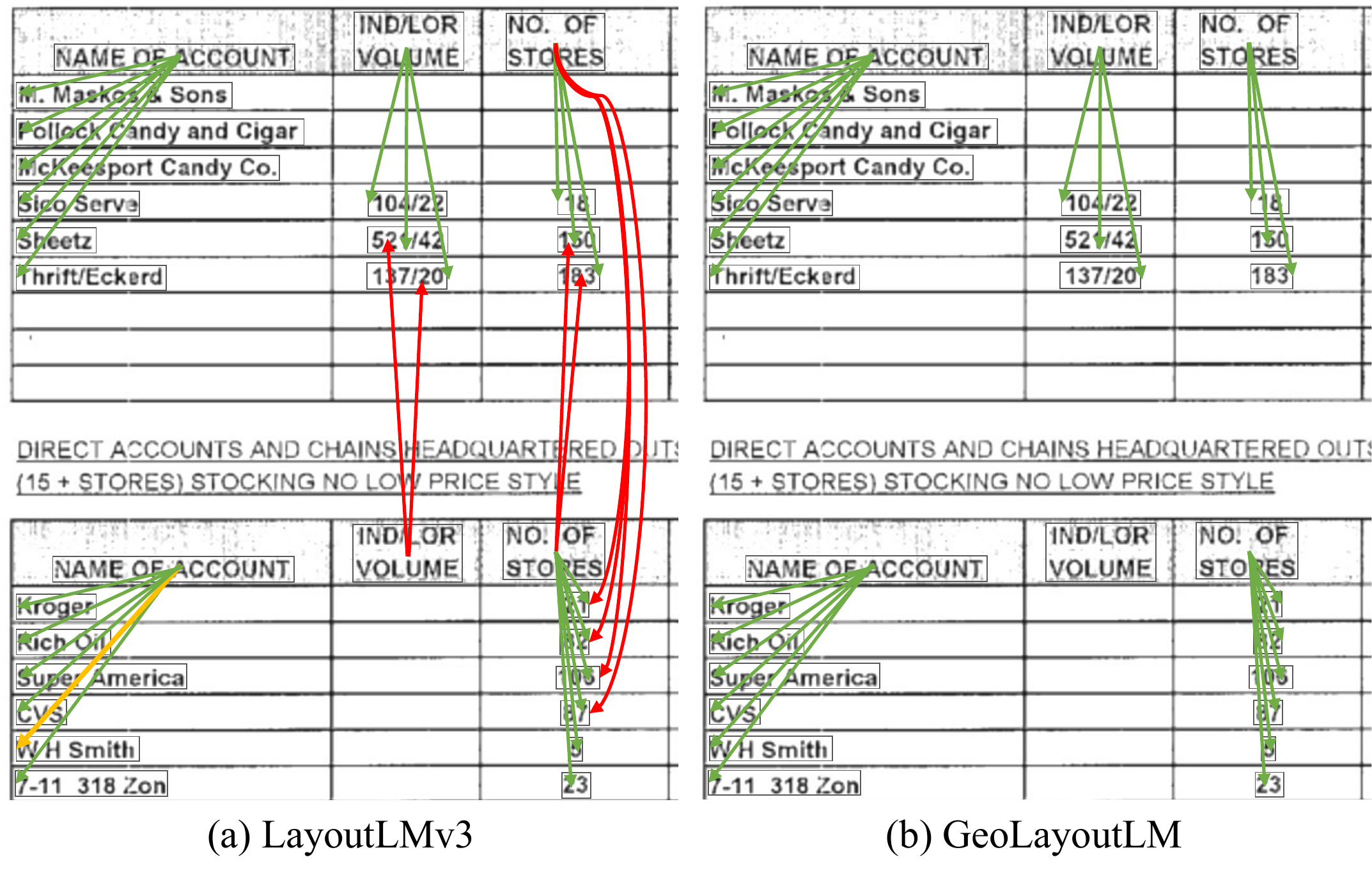}
   \vspace{-2mm}
   \caption{RE case study. The arrows in green, red and orange denote true positive, false positive and false negative (missed) relations respectively. Best viewed by zooming up.}
   \label{link_case}
\end{figure}

A case study is given in \cref{link_case}. Most of the false positive relation links predicted by LayoutLMv3 violate the geometric layout obviously. It depends on the semantic information excessively and ignores the layout more or less. For example, the entity starting with ``No.'' is linked to the number entity regardless of the geometric relationship between them. In contrast, GeoLayoutLM successfully predicted all links with a good recall. For more details, please refer to the appendix\ref{sec:case_study}.

Although the rule-based geometric constraint can bring some improvement (\cref{intro_v3_geo}), it still fall behind GeoLayoutLM because it:
(1) relies on hard-coded thresholds, which limit its adaptability and generalization when handling documents of different formats and layouts; (2) is able to prune false linkings, but cannot recover missed ones.

% \subsubsection{Effects of the Relation Heads}
\noindent \textbf{Effects of the Relation Heads.}
There are two important points in our RE task heads: the novel RFE head and its pre-training. We study the impact of them for the RE task. The coarse relation prediction (CRP) head is always kept. Besides, we do not use the RSF strategy to be elegant.

As shown in \cref{abl_link_head}, a bare CRP head not initialized by the pre-trained parameters (w/o Pt) achieves an 82.2\% F1 score owing to the strong geometry-aware backbone. Once it is initialized by pre-training (Pt), an improvement of 2.7\% F1 score is obtained. By adding the RFE head (w/o Pt), the version of CRP (w/o Pt) becomes stronger while that of CRP (Pt) even degrades a little bit. We argue that the RFE head introduces more parameters, which causes overfitting despite its superiority in relation modeling. Thus it is necessary to pre-train the RFE head. We also find that the pre-training of the RFE head is more important than that of the CRP head. By making full use of the two points, GeoLayoutLM obtains the best RE performance.

\begin{table}
  \centering
  \begin{tabular}{m{.14\columnwidth}<{\centering}m{.14\columnwidth}<{\centering}|m{.14\columnwidth}<{\centering}m{.14\columnwidth}<{\centering}|m{.12\columnwidth}<{\centering}}
    \toprule[1pt]
    \multicolumn{2}{c|}{\textbf{CRP head}} & \multicolumn{2}{c|}{\textbf{RFE head}} & \multirow{2}*{\textbf{F1}}\\
    \cline{1-4}
    w/o Pt & Pt & w/o Pt & Pt & \\
    \hline
    \checkmark & & & & 82.2\\
     & \checkmark &&& 84.9\\
    \hline
    \checkmark &&\checkmark&& 84.0\\
    & \checkmark & \checkmark && 84.0\\
    \checkmark &&&\checkmark& 86.2\\
    & \checkmark &&\checkmark& \textbf{86.9}\\
    \bottomrule[1pt]
  \end{tabular}
  \caption{Ablation study on the CRP and RFE head in FUNSD RE task. ``w/o Pt" and ``Pt" mean that the head is not pre-trained and pre-trained respectively. The RSF strategy is not used.}
  \label{abl_link_head}
  \vspace{-4mm}
\end{table}

% \begin{table}
%   \centering
%   \begin{tabular}{m{.17\columnwidth}<{\centering}|m{.17\columnwidth}<{\centering}m{.17\columnwidth}<{\centering}}
%     \toprule[1pt]
%     \multirow{2}*{Memory} & FUNSD & CORD\\
%     \cline{2-3}
%     & RE & REaGRP\\
%     \hline
%     Frozen & 88.17 & \\
%     Encoded & 89.45 & 99.45\\
%     \bottomrule[1pt]
%   \end{tabular}
%   \caption{Ablation study of the encoder in RFE head.}
%   \label{abl_rfe}
% \end{table}

% \subsubsection{Effects of RSF}
\noindent \textbf{Effects of RSF.}
The RSF strategy is non-trivial in the fine-tuning and inference stage. It contains two parts: the post-process for inference and the variance loss in fine-tuning.

\cref{abl_rsf} gives a clear view. By the post-processing, the precision of our method is improved dramatically with a little sacrifice of recall. A bare variance loss without the post-processing does nothing to the performance since it is designed for the post-processing only. We obtain the best F1 score when using both of them.

\begin{table}
  \centering
  \begin{tabular}{m{.16\columnwidth}<{\centering}m{.15\columnwidth}<{\centering}|m{.14\columnwidth}<{\centering}m{.11\columnwidth}<{\centering}m{.10\columnwidth}<{\centering}}
    \toprule[1pt]
    \textbf{postprocess} &\textbf{variance loss} & \textbf{Precision} & \textbf{Recall} & \textbf{F1}\\
    \hline
    && 85.26 & 90.15 & 87.64\\
    \checkmark && 88.25 & 89.01 & 88.62\\
    & \checkmark & 85.06 & 90.34 & 87.62\\
    \checkmark & \checkmark & 88.94 & 89.96 & 89.45\\
    \bottomrule[1pt]
  \end{tabular}
  \caption{Ablation study on the RSF strategy in FUNSD RE task.}
  \label{abl_rsf}
  \vspace{-2mm}
\end{table}

\subsection{Few-shot RE Learning}

In real scenarios, the acquirement of the training data for document information extraction is a bottleneck due to the expensive and boring annotation work. So it is necessary to learn from only a few document samples.

To explore the ability of few-shot learning, we compare our GeoLayoutLM with another two models: a modified GeoLayoutLM whose heads are not initialized from pre-training (GeoLayoutLM*), and LayoutLMv3. We also disable the RSF strategy to make it clearer.

\begin{figure}[t]
  \centering
   \includegraphics[width=0.9\linewidth]{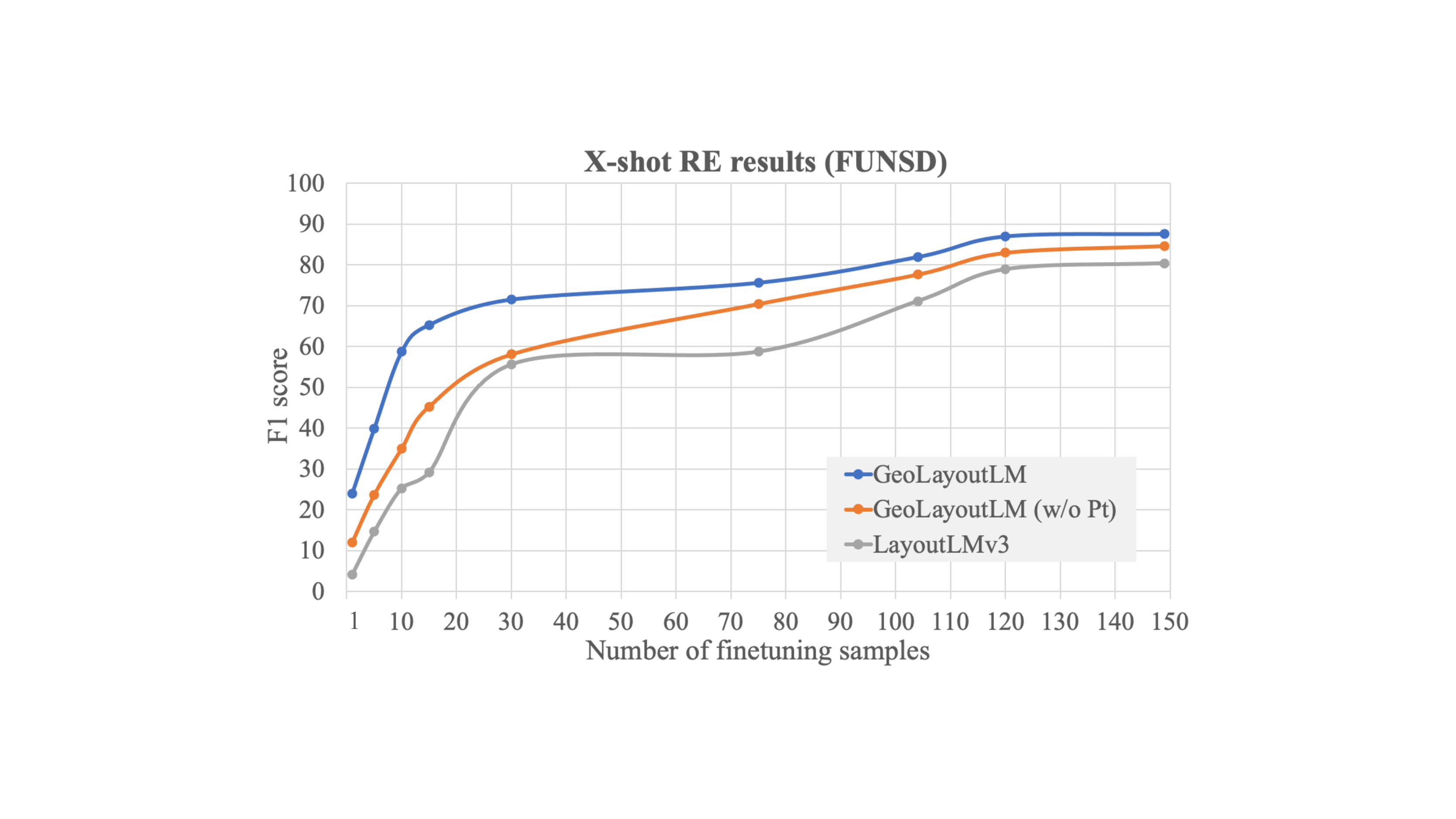}
   \caption{Comparison of few-shot learning in FUNSD RE task.}
   \label{fewshot}
   \vspace{-3mm}
\end{figure}

As shown in \cref{fewshot}, GeoLayoutLM shows great superiority in this setting. GeoLayoutLM* outperforms LayoutLMv3 but is inferior to GeoLayoutLM all the time. It suggests that the geometric pre-training endows our model with a strong ability to extract entity relations, and also emphasizes the importance of RE head pre-training. Notably, our GeoLayoutLM achieves a slightly better performance (71.53\%) using only 30 samples than LayoutLMv3 does (71.07\%) using 104 samples. The performance gap is very large when only few fine-tuning samples are available.

% \subsection{Zero-shot transferring in multi-lingual task}

\section{Conclusion}

In this paper, we propose GeoLayoutLM, a geometric pre-training framework for VIE.
Three geometric relations in different levels are defined: GeoPair, GeoMPair and GeoTriplet.
Correspondingly, three specially designed pre-training objectives are introduced to model geometric relations explicitly.
Additionally, the relation heads are elaborately designed to enhance the relation feature representation, which are pre-trained by the geometric pre-training, thus mitigating the gap between pre-training and fine-tuning.
Experimental results on VIE have illustrated the effectiveness of GeoLayoutLM in both SER and RE tasks.
% Moreover, the ablation studies demonstrate that GeoLayoutLM is able to understand the geometric layout information in documents.
In the future, we will explore more effective geometric pre-training tasks, and apply our method to more tasks of visually-rich document understanding.

%%%%%%%%% REFERENCES
{\small
\bibliographystyle{ieee_fullname}
\bibliography{egbib}

\begin{thebibliography}{10}\itemsep=-1pt

\bibitem{appalaraju2021docformer}
Srikar Appalaraju, Bhavan Jasani, and Bhargava~Urala Kota.
\newblock {DocFormer}: End-to-end transformer for document understanding.
\newblock In {\em ICCV}, pages 4171--4186, 2021.

\bibitem{chronopoulou2019embarrassingly}
Alexandra Chronopoulou, Christos Baziotis, and Alexandros Potamianos.
\newblock An embarrassingly simple approach for transfer learning from
  pretrained language models.
\newblock {\em NAACL}, 2019.

\bibitem{cui2021document}
Lei Cui, Yiheng Xu, Tengchao Lv, and Furu Wei.
\newblock Document ai: Benchmarks, models and applications.
\newblock {\em arXiv preprint arXiv:2111.08609}, 2021.

\bibitem{devlin2018bert}
Jacob Devlin, Ming-Wei Chang, Kenton Lee, and Kristina Toutanova.
\newblock Bert: Pre-training of deep bidirectional transformers for language
  understanding.
\newblock {\em arXiv preprint arXiv:1810.04805}, 2018.

\bibitem{dosovitskiy2020image}
Alexey Dosovitskiy, Lucas Beyer, Alexander Kolesnikov, Dirk Weissenborn,
  Xiaohua Zhai, Thomas Unterthiner, Mostafa Dehghani, Matthias Minderer, Georg
  Heigold, Sylvain Gelly, et~al.
\newblock An image is worth 16x16 words: Transformers for image recognition at
  scale.
\newblock {\em arXiv preprint arXiv:2010.11929}, 2020.

\bibitem{dou2022empirical}
Zi-Yi Dou, Yichong Xu, Zhe Gan, Jianfeng Wang, Shuohang Wang, Lijuan Wang,
  Chenguang Zhu, Pengchuan Zhang, Lu Yuan, Nanyun Peng, et~al.
\newblock An empirical study of training end-to-end vision-and-language
  transformers.
\newblock In {\em CVPR}, pages 18166--18176, 2022.

\bibitem{gu2022unified}
Jiuxiang Gu, Jason Kuen, Vlad~I Morariu, Handong Zhao, Nikolaos Barmpalios,
  Rajiv Jain, Ani Nenkova, and Tong Sun.
\newblock Unified pretraining framework for document understanding.
\newblock In {\em NeurIPS}, 2021.

\bibitem{gu2022xylayoutlm}
Zhangxuan Gu, Changhua Meng, Ke Wang, Jun Lan, Weiqiang Wang, Ming Gu, and
  Liqing Zhang.
\newblock Xylayoutlm: Towards layout-aware multimodal networks for
  visually-rich document understanding.
\newblock In {\em CVPR}, pages 4583--4592, 2022.

\bibitem{gururangan2020don}
Suchin Gururangan, Ana Marasovi'c, Swabha Swayamdipta, Kyle Lo, Iz Beltagy,
  Doug Downey, and Noah~A Smith.
\newblock Don't stop pretraining: adapt language models to domains and tasks.
\newblock {\em ACL}, 2020.

\bibitem{han2021adaptive}
Xueting Han, Zhenhuan Huang, Bang An, and Jing Bai.
\newblock Adaptive transfer learning on graph neural networks.
\newblock In {\em SIGKDD}, 2021.

\bibitem{he2017mask}
Kaiming He, Georgia Gkioxari, Piotr Doll{\'a}r, and Ross Girshick.
\newblock Mask r-cnn.
\newblock In {\em ICCV}, pages 2961--2969, 2017.

\bibitem{hong2022bros}
Teakgyu Hong, Donghyun Kim, Mingi Ji, Wonseok Hwang, Daehyun Nam, and Sungrae
  Park.
\newblock Bros: A pre-trained language model focusing on text and layout for
  better key information extraction from documents.
\newblock In {\em AAAI}, 2022.

\bibitem{howard2018universal}
Jeremy Howard and Sebastian Ruder.
\newblock Universal language model fine-tuning for text classification.
\newblock {\em ACL}, 2018.

\bibitem{hu2022p}
Xiaomeng Hu, Shi Yu, Chenyan Xiong, Zhenghao Liu, Zhiyuan Liu, and Ge Yu.
\newblock P $^3$ ranker: Mitigating the gaps between pre-training and ranking
  fine-tuning with prompt-based learning and pre-finetuning.
\newblock {\em SIGIR}, 2022.

\bibitem{huang2022layoutlmv3}
Yupan Huang, Tengchao Lv, Lei Cui, Yutong Lu, and Furu Wei.
\newblock Layoutlmv3: Pre-training for document ai with unified text and image
  masking.
\newblock In {\em ACM Multimedia}, 2022.

\bibitem{hwang2020spatial}
Wonseok Hwang, Jinyeong Yim, Seunghyun Park, Sohee Yang, and Minjoon Seo.
\newblock Spatial dependency parsing for semi-structured document information
  extraction.
\newblock {\em ACL findings}, 2021.

\bibitem{jaume2019funsd}
Guillaume Jaume, Hazim~Kemal Ekenel, and Jean-Philippe Thiran.
\newblock {Funsd}: A dataset for form understanding in noisy scanned documents.
\newblock In {\em ICDARW}, volume~2, pages 1--6, 2019.

\bibitem{kipf2016semi}
Thomas~N Kipf and Max Welling.
\newblock Semi-supervised classification with graph convolutional networks.
\newblock In {\em ICLR}, 2017.

\bibitem{lewis2006building}
David Lewis, Gady Agam, Shlomo Argamon, Ophir Frieder, David Grossman, and
  Jefferson Heard.
\newblock Building a test collection for complex document information
  processing.
\newblock In {\em SIGIR}, 2006.

\bibitem{li2021structurallm}
Chenliang Li, Bin Bi, and Ming Yan.
\newblock {StructuralLM}: Structural pre-training for form understanding.
\newblock In {\em ACL}, 2021.

\bibitem{li2021selfdoc}
Peizhao Li, Jiuxiang Gu, and Jason Kuen.
\newblock {SelfDoc}: Self-supervised document representation learning.
\newblock In {\em CVPR}, pages 5652--5660, 2021.

\bibitem{li2021structext}
Yulin Li, Yuxi Qian, Yuechen Yu, Xiameng Qin, Chengquan Zhang, Yan Liu, Kun
  Yao, Junyu Han, Jingtuo Liu, and Errui Ding.
\newblock Structext: Structured text understanding with multi-modal
  transformers.
\newblock In {\em ACM Multimedia}, pages 1912--1920, 2021.

\bibitem{liao2020real}
Minghui Liao, Zhaoyi Wan, Cong Yao, Kai Chen, and Xiang Bai.
\newblock Real-time scene text detection with differentiable binarization.
\newblock In {\em AAAI}, volume~34, pages 11474--11481, 2020.

\bibitem{lin2013network}
Min Lin, Qiang Chen, and Shuicheng Yan.
\newblock Network in network.
\newblock In {\em ICLR}, 2014.

\bibitem{lin2021vibertgrid}
Weihong Lin, Qifang Gao, Lei Sun, Zhuoyao Zhong, Kai Hu, Qin Ren, and Qiang
  Huo.
\newblock Vibertgrid: a jointly trained multi-modal 2d document representation
  for key information extraction from documents.
\newblock In {\em ICDAR}, pages 548--563. Springer, 2021.

\bibitem{liu2022prompt}
Pengfei Liu, Weizhe Yuan, Jinlan Fu, Zhengbao Jiang, Hiroaki Hayashi, and
  Graham Neubig.
\newblock Pre-train, prompt, and predict: A systematic survey of prompting
  methods in natural language processing.
\newblock {\em ACM Comput. Surv.}, sep 2022.
\newblock Just Accepted.

\bibitem{liu2019graph}
Xiaojing Liu, Feiyu Gao, and Qiong Zhang.
\newblock Graph convolution for multimodal information extraction from visually
  rich documents.
\newblock In {\em NAACL-HLT (1)}, pages 32--39, 2019.

\bibitem{liu2022convnet}
Zhuang Liu, Hanzi Mao, Chao-Yuan Wu, Christoph Feichtenhofer, Trevor Darrell,
  and Saining Xie.
\newblock A convnet for the 2020s.
\newblock In {\em CVPR}, pages 11976--11986, 2022.

\bibitem{Long2018SceneTD}
Shangbang Long, Xin He, and Cong Yao.
\newblock Scene text detection and recognition: The deep learning era.
\newblock {\em International Journal of Computer Vision}, 129:161--184, 2018.

\bibitem{luo2022bivldoc}
Chuwei Luo, Guozhi Tang, Qi Zheng, Cong Yao, Lianwen Jin, Chenliang Li, Yang
  Xue, and Luo Si.
\newblock Bi-vldoc: Bidirectional vision-language modeling for visually-rich
  document understanding.
\newblock {\em arXiv preprint arXiv:2206.13155}, 2022.

\bibitem{luo2020merge}
Chuwei Luo, Yongpan Wang, Qi Zheng, Liangchen Li, Feiyu Gao, and Shiyu Zhang.
\newblock Merge and recognize: a geometry and 2d context aware graph model for
  named entity recognition from visual documents.
\newblock In {\em Proceedings of the Graph-based Methods for Natural Language
  Processing (TextGraphs)}, pages 24--34, 2020.

\bibitem{park2019cord}
Seunghyun Park, Seung Shin, Bado Lee, and Junyeop Lee.
\newblock {CORD}: A consolidated receipt dataset for post-ocr parsing.
\newblock In {\em Document Intelligence Workshop at NeurIPS}, 2019.

\bibitem{qian2019graphie}
Yujie Qian, Enrico Santus, and Zhijing Jin.
\newblock {GraphIE}: A graph-based framework for information extraction.
\newblock In {\em NAACL}, pages 751--761, 2019.

\bibitem{qian2018graphie}
Yujie Qian, Enrico Santus, Zhijing Jin, Jiang Guo, and Regina Barzilay.
\newblock Graphie: A graph-based framework for information extraction.
\newblock In {\em NAACL}, 2019.

\bibitem{Shi2015AnET}
Baoguang Shi, Xiang Bai, and Cong Yao.
\newblock An end-to-end trainable neural network for image-based sequence
  recognition and its application to scene text recognition.
\newblock {\em IEEE Transactions on Pattern Analysis and Machine Intelligence},
  39:2298--2304, 2015.

\bibitem{tang2021matchvie}
Guozhi Tang, Lele Xie, Lianwen Jin, and Wang.
\newblock {MatchVIE}: Exploiting match relevancy between entities for visual
  information extraction.
\newblock In {\em IJCAI}, 2021.

\bibitem{vaswani2017attention}
Ashish Vaswani, Noam Shazeer, Niki Parmar, Jakob Uszkoreit, Llion Jones,
  Aidan~N Gomez, {\L}ukasz Kaiser, and Illia Polosukhin.
\newblock Attention is all you need.
\newblock {\em Advances in neural information processing systems}, 30, 2017.

\bibitem{wang2022lilt}
Jiapeng Wang, Lianwen Jin, and Kai Ding.
\newblock Lilt: A simple yet effective language-independent layout transformer
  for structured document understanding.
\newblock In {\em ACL}, 2022.

\bibitem{wang2022multi}
Peng Wang, Cheng Da, and Cong Yao.
\newblock Multi-granularity prediction for scene text recognition.
\newblock In {\em Computer Vision--ECCV 2022: 17th European Conference, Tel
  Aviv, Israel, October 23--27, 2022, Proceedings, Part XXVIII}, pages
  339--355. Springer, 2022.

\bibitem{xu2020layoutlm}
Yiheng Xu, Minghao Li, Lei Cui, and Shaohan Huang.
\newblock {LayoutLM}: Pre-training of text and layout for document image
  understanding.
\newblock In {\em KDD}, pages 1192--1200, 2020.

\bibitem{xu2021layoutxlm}
Yiheng Xu, Tengchao Lv, Lei Cui, Guoxin Wang, Yijuan Lu, Dinei Florencio, Cha
  Zhang, and Furu Wei.
\newblock Layoutxlm: Multimodal pre-training for multilingual visually-rich
  document understanding.
\newblock {\em arXiv preprint arXiv:2104.08836}, 2021.

\bibitem{xu2022xfund}
Yiheng Xu, Tengchao Lv, Lei Cui, Guoxin Wang, Yijuan Lu, Dinei Florencio, Cha
  Zhang, and Furu Wei.
\newblock Xfund: A benchmark dataset for multilingual visually rich form
  understanding.
\newblock In {\em ACL Findings}, pages 3214--3224, 2022.

\bibitem{xu2020layoutlmv2}
Yang Xu, Yiheng Xu, Tengchao Lv, Lei Cui, Furu Wei, Guoxin Wang, Yijuan Lu,
  Dinei Florencio, Cha Zhang, Wanxiang Che, Min Zhang, and Lidong Zhou.
\newblock Layoutlmv2: Multi-modal pre-training for visually-rich document
  understanding.
\newblock In {\em ACL}, 2021.

\bibitem{yu2021pick}
Wenwen Yu, Ning Lu, Xianbiao Qi, Ping Gong, and Rong Xiao.
\newblock Pick: Processing key information extraction from documents using
  improved graph learning-convolutional networks.
\newblock In {\em ICPR}, 2020.

\bibitem{zhang2021entity}
Yue Zhang, Bo Zhang, Rui Wang, Junjie Cao, Chen Li, and Zuyi Bao.
\newblock Entity relation extraction as dependency parsing in visually rich
  documents.
\newblock {\em EMNLP}, 2021.

\bibitem{Zhou2017EASTAE}
Xinyu Zhou, Cong Yao, He Wen, Yuzhi Wang, Shuchang Zhou, Weiran He, and Jiajun
  Liang.
\newblock East: An efficient and accurate scene text detector.
\newblock {\em 2017 IEEE Conference on Computer Vision and Pattern Recognition
  (CVPR)}, pages 2642--2651, 2017.

\bibitem{Zhu2016SceneTD}
Yingying Zhu, Cong Yao, and Xiang Bai.
\newblock Scene text detection and recognition: recent advances and future
  trends.
\newblock {\em Frontiers of Computer Science}, 10:19--36, 2016.

\end{thebibliography}
}

\newpage
\appendix
\section{Appendix}

\subsection{Case Study}
\label{sec:case_study}

\begin{figure*}[tp]
  \centering
  \includegraphics[width=1.0\linewidth]{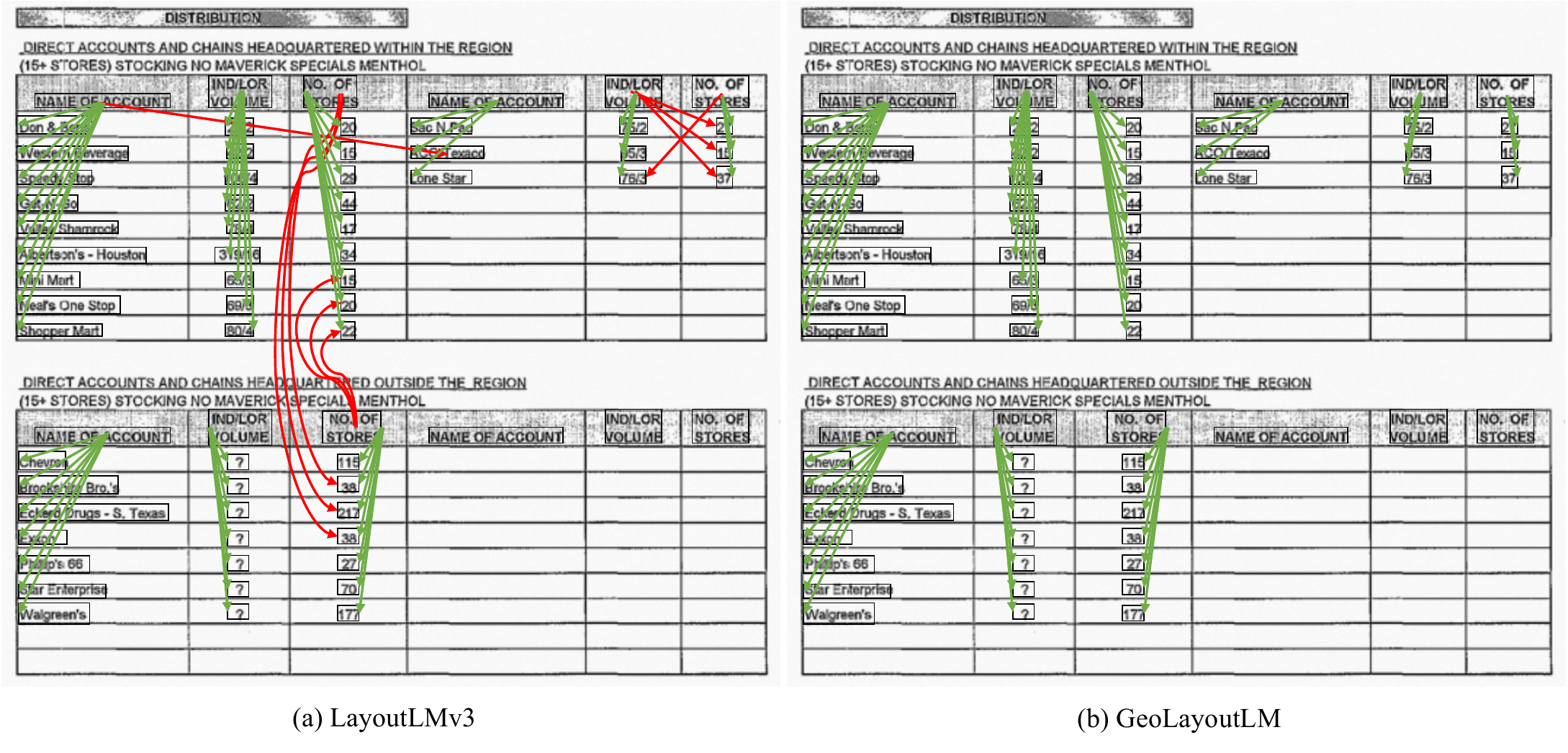}
  \caption{RE case study in tables.}
  \label{appendix_case_study_1}
\end{figure*}

\begin{figure*}[tp]
  \centering
  \includegraphics[width=1.0\linewidth]{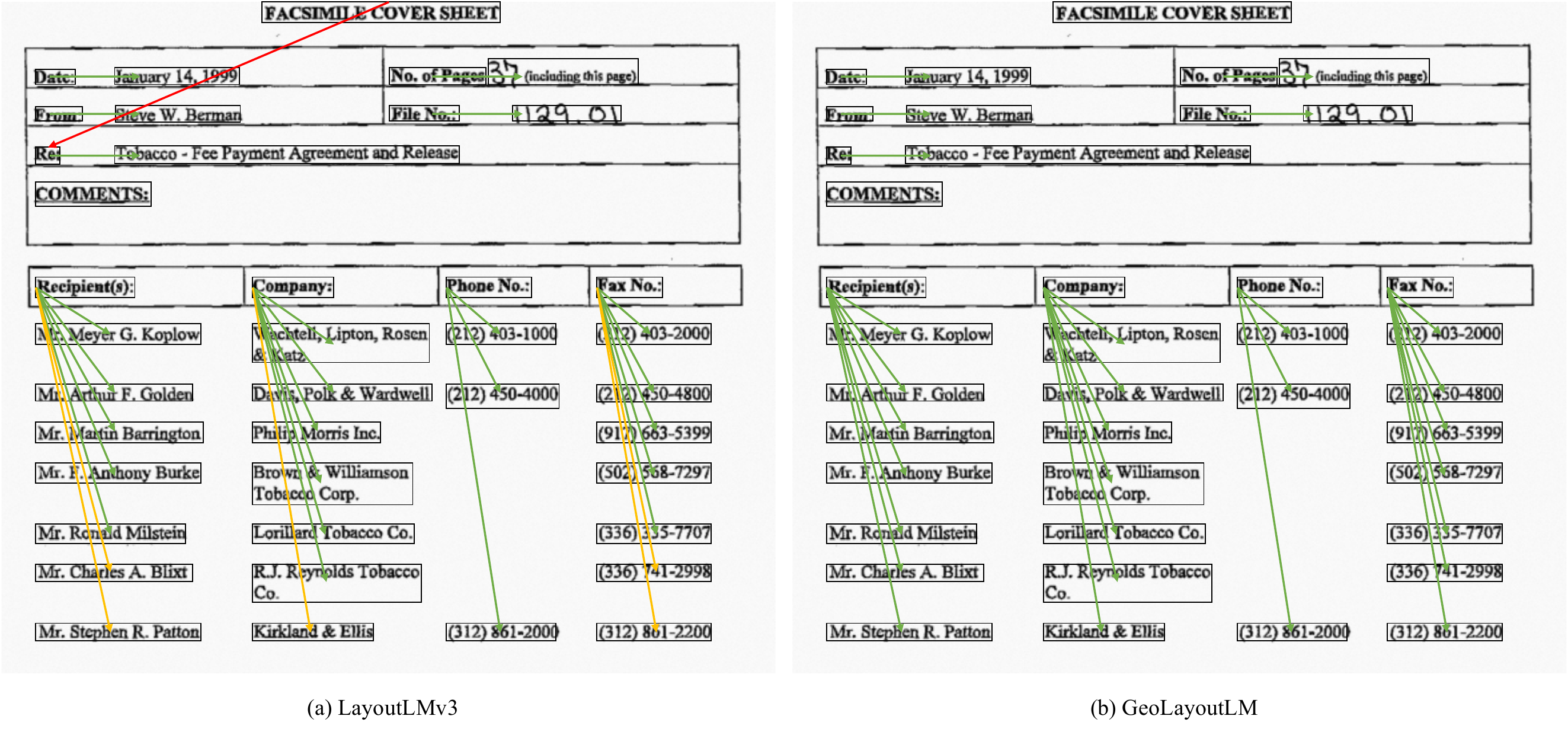}
  \caption{RE case study in a borderless table.}
  \label{appendix_case_study_2}
\end{figure*}

\begin{figure*}[tp]
  \centering
  \includegraphics[width=1.0\linewidth]{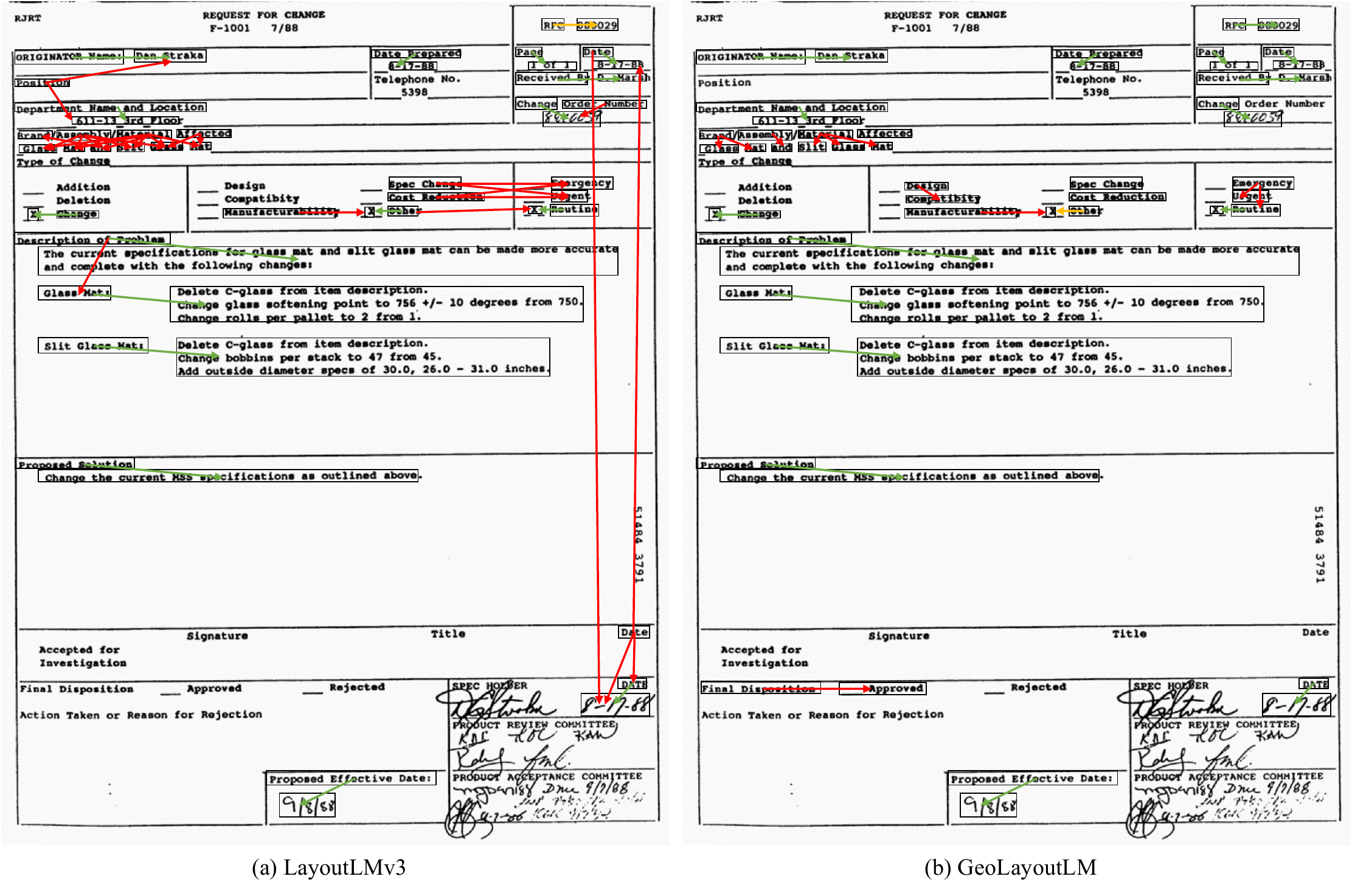}
  \caption{Failure cases comparison.}
  \label{appendix_case_study_3}
\end{figure*}

One motivation behind GeoLayoutLM is to avoid making predictions that violate the geometric layout.
As \cref{appendix_case_study_1} shows, GeoLayoutLM successfully predicts all links in a document that contains multiple tables. 
However, LayoutLMv3 tends to link two entities relying more on their semantics.
Most of the false positive relation links predicted by LayoutLMv3 are cross tables or table columns, which is counter-intuitive in terms of the geometric layout.
Another motivation behind GeoLayoutLM is to predicts right links that are similar in geometric layout.
A case study is given in \cref{appendix_case_study_2}.
In this case, LayoutLMv3 successfully predicts the link in the upper half part of the wireless form but misses the link below. For example, the ``Recipient(s)'' is linked to ``Mr. Ronald Milstein'' and ``Mr. F. Anthony Burke'', but the ``Recipient(s)'' misses ``Mr. Charles A. Blixt'' and ``Mr. Stephen R. Patton'', despite that these links are similar in geometric layout.
Our GeoLayoutLM extracts all the relations exactly.
A failure example is given in \cref{appendix_case_study_3}. 
Most of the false positive relation links predicted by GeoLayoutLM still conforms to the geometric layout. There are no false positive relation links that are cross the borderline of the table.
These examples show that GeoLayoutLM predicts relation links more accurately, and complies with the geometric layout restriction well.

\end{document}